%% file: main.tex
\documentclass[10pt,twocolumn,letterpaper]{article}

\usepackage[pagenumbers]{cvpr}

\input{preamble}

\definecolor{cvprblue}{rgb}{0.21,0.49,0.74}
\usepackage[pagebackref,breaklinks,colorlinks,allcolors=cvprblue]{hyperref}

\title{Wizmos: Parameterized Word Gizmos for Multimodal Image Attribute Control}

\author{
Qingcheng Zhao$^{1}$ \qquad Karran Pandey$^{1}$ \qquad Or Patashnik$^{2}$ \qquad Karan Singh$^{1}$\\[0.5em]
$^{1}$University of Toronto \qquad $^{2}$Tel Aviv University
}

\begin{document}
\maketitle
\input{fig/teaser}

\input{sec/0_abstract}
\input{sec/1_intro}

\input{sec/2_related}

\input{sec/3_method}

\input{sec/4_experiments}

\input{sec/5_conclusion}

{
    \small
    \bibliographystyle{ieeenat_fullname}
    \bibliography{main}
}
\input{sec/A_appendix}

\end{document}

%% file: preamble.tex
\usepackage{cuted}

\usepackage{currfile}

\usepackage{caption}

\usepackage{graphicx}
\usepackage{amsmath,amssymb}
\usepackage{booktabs}
\usepackage{xspace}
\usepackage{multirow}
\newcommand{\klein}{FLUX.2~[klein]\xspace}

\usepackage{pifont}

\newcommand{\cmark}{\ding{51}}
\newcommand{\xmark}{\ding{55}}
\DeclareRobustCommand{\pmark}{\tikz[baseline=-0.8ex]{\draw[line width=0.4pt] (0,0) circle (0.8ex); \fill (0,0) -- (90:0.8ex) arc (90:270:0.8ex) -- cycle;}}
\newcommand{\na}{--}
\newcommand{\up}{$\uparrow$}
\newcommand{\down}{$\downarrow$}
\usepackage{colortbl}
\usepackage{adjustbox}
\usepackage{pgfplots}
\pgfplotsset{compat=1.16}
\usepgfplotslibrary{groupplots}
\newcommand{\tgroup}[2]{\rowcolor{gray!12}\multicolumn{#1}{@{}l}{\textbf{#2}}\\}
\definecolor{tbest}{RGB}{161,217,155}
\definecolor{tsecond}{RGB}{222,242,217}
\newcommand{\best}{\cellcolor{tbest}}
\newcommand{\second}{\cellcolor{tsecond}}
\definecolor{tworst}{RGB}{244,165,155}
\definecolor{tsworst}{RGB}{252,224,218}
\newcommand{\worst}{\cellcolor{tworst}}
\newcommand{\sworst}{\cellcolor{tsworst}}

%% file: fig/teaser.tex
\begin{strip}
\centering
\includegraphics[width=\textwidth]{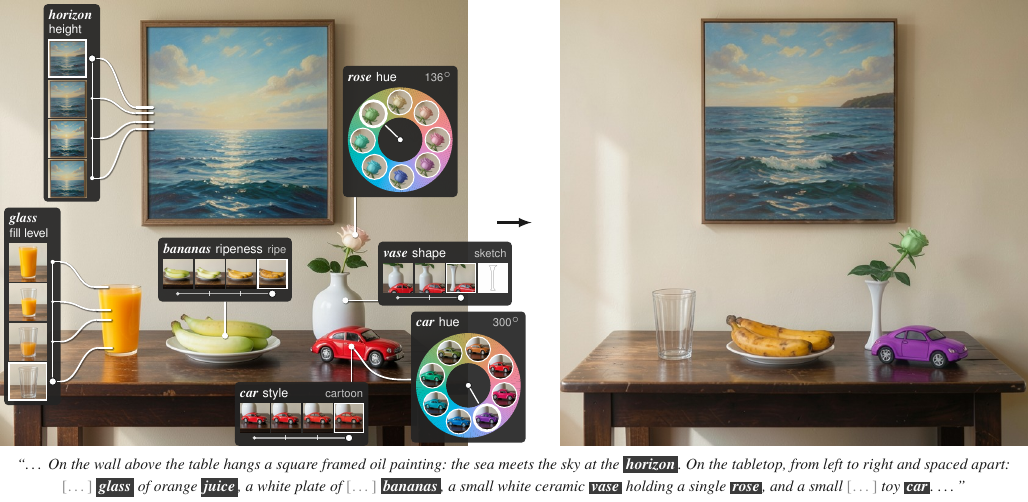}
\captionsetup{hypcap=false}
\captionof{figure}{Wizmos: reusable, parameterized controls
attached to prompt words. Each wizmo maps parameters in a
continuous attribute domain to visual changes
in what the word describes. Here, six wizmos vary multimodal
attributes: horizon height, glass fill level, banana ripeness,
vase shape, rose hue and toy car style. Hue and style compose
on the toy car, while all controls act together across the
scene. Results are shown at selected parameter values,
with insets illustrating intermediates.}
\captionsetup{hypcap=true}
\label{fig:teaser}%
\end{strip}

%% file: sec/0_abstract.tex
\begin{abstract}
Input prompts to text-to-image models leave many visual choices unspecified,
with few ways to explore them. We introduce \emph{wizmos}
(word gizmos): reusable, parameterized controls attached
to prompt words, mapping meaningful attribute parameters
to corresponding visual changes. Each wizmo exposes
a continuous attribute-specific domain, such as a colour wheel or a two-dimensional spatial region, through which users
can explore different values of the attribute.
Once constructed, wizmos transfer across generations and
compose on and across objects, forming a reusable
toolbox for exploring the generator's unspecified choices.
Our general framework learns mappings from attribute parameters to
contextualized text-state offsets in a frozen multimodal
diffusion transformer. We fit these mappings to produce continuous visual changes along attribute scales defined through geometric measurements, reference comparisons, or scores from predictors and
VLMs.
We demonstrate wizmos spanning geometry, appearance,
and semantic and perceptual attributes, including
specifications and parameterizations unsupported by
prior art. We evaluate their accuracy, progression,
preservation, transfer and composition, and show
applications in interactive editing and exploration.
\par\vspace{0.5em}\centerline{\small Project page: \href{https://www.cs.toronto.edu/~zhaoqi/wizmos/}{\nolinkurl{www.cs.toronto.edu/~zhaoqi/wizmos}}}
\end{abstract}

%% file: sec/1_intro.tex
\section{Introduction}
\label{sec:intro}

Text-to-image models produce detailed, photorealistic images from short prompts, but make many visual choices on the user's behalf.
A prompt can name a glass of juice, a rose in a vase and a seascape
on the wall without fixing how full the glass is, how high the
horizon sits, what color the rose is or what shape the vase takes
(\cref{fig:teaser}).
Some details are easy to describe; others are easier to show. Many vary continuously, leaving possibilities to explore. These choices form the prompt's \emph{null space}: variation that remains faithful to its textual description.

Although users may find it difficult to specify exactly what they want from this space, they can often identify attributes they would like to explore. Visual attributes span a multimodal spectrum: from
geometric properties, such as the vase's shape, through appearance
properties, such as the rose's color and the toy car's style, to
semantic and perceptual properties, such as the bananas'
ripeness~\cite{brath2013visual, isola2015discovering,
parikh2011relative}.

We introduce \emph{wizmos} (word gizmos\footnote{In graphics
interfaces, a gizmo is an interactive control for manipulating
an object, such as a handle for translation, rotation or scaling.}): reusable, parameterized
controls that can be attached to prompt words to explore its attributes. Each wizmo exposes a
continuous attribute domain for a selected prompt word, mapping
parameters in that domain to corresponding visual changes in what the
word describes. A hue wizmo on ``rose'', for example, lets users
explore its color by selecting angles around a color wheel. Once
constructed, wizmos can be applied to other applicable words, reused
across prompts and seeds, and combined to control multiple
attributes in the same image, forming a reusable
toolbox for exploring a prompt's null space.

Text conditioning offers a natural way to realize wizmos. Prior work
steers generation through offsets to prompt
representations~\cite{baumann2025continuous, ekin2026unreasonable,
kamenetsky2025saedit}, drawing on the generator's learned visual
prior without modifying its weights. These offsets can be reused and
combined, with their strengths serving as continuous control
settings. However, language-defined or discovered directions and
their strengths may not provide the right abstraction for users: the
vase's outline or toy car's style may be better conveyed by a
sketch or reference, while exploring a rose's hue calls for access
to the color wheel, not simply making it more or less green.

\input{tab/related_work}

We present a general framework for constructing wizmos from visual
evaluations. Our key observation is that visual variation can often
be evaluated even when it is difficult to specify in words or
demonstrate with paired examples. Geometry can be measured directly,
appearance compared to visual references, and semantic or perceptual
properties assessed by predictors and VLMs.

Technically, we realize each wizmo as a learned continuous
mapping from attribute parameter values to offsets in a
prompt word's contextualized text states. We use a frozen
multimodal diffusion transformer (MMDiT), where word
representations interact with image representations and
acquire information about the emerging
scene~\cite{dahary2026contextual}.
We fit the mapping across source generations and parameter
values sampled throughout the attribute domain, using
visual evaluations to learn a consistent attribute
response across its scale.
The mapping predicts offsets at intermediate values
not sampled during fitting, allowing users to explore
different trajectories through the domain (\cref{fig:sliders}). It can be applied to other applicable prompt words and real images without modifying the generator (\cref{fig:gallery}).

We demonstrate wizmos spanning geometry, appearance, and semantic
and perceptual properties (\cref{fig:gallery}), including specifications and
parameterizations unsupported by prior art
(\cref{fig:sliders}). Together, they
form a reusable toolbox for interactive editing and user-defined
exploration of the generator (\cref{fig:applications}). Our contributions are:
\begin{enumerate}
    \item \textbf{Parameterized controls attached to prompt
    words.}
    We introduce wizmos: reusable, extendable tools that expose an attribute domain for a selected prompt word, mapping its parameters to visual changes. We present a general framework
    for constructing wizmos across the spectrum of visual
    attributes using visual evaluations (\cref{fig:method,fig:gallery}).
    \item \textbf{Meaningful attribute parameterizations.}
    We tie wizmos to attribute parameters over continuous domains
    suited to each attribute, including cyclic and multidimensional spaces (\cref{fig:sliders}).
    \item \textbf{Reuse and composition through the generator's
    prior.}
    We show that contextualized prompt-word offsets support
    localized attribute changes that transfer across generations and combine within and across objects (\cref{fig:teaser}).
\end{enumerate}

%% file: tab/related_work.tex
\begin{table}[t]
\centering
\footnotesize
\setlength{\tabcolsep}{4pt}
\begin{adjustbox}{max width=\linewidth}
\begin{tabular}{@{}lccc@{}}
\toprule
Method
& \shortstack{Multimodal}
& \shortstack{Continuous}
& \shortstack{Attribute Parameterization} \\
\midrule
Text-space steering~\cite{baumann2025continuous,ekin2026unreasonable,kamenetsky2025saedit}
& \xmark & \cmark & \xmark \\
Concept Sliders~\cite{gandikota2024concept}
& \cmark & \cmark & \xmark \\
Dual-Process~\cite{luo2025dual}
& \cmark & \pmark & \xmark \\
\midrule
SliderEdit~\cite{zarei2026slideredit}
& \xmark & \cmark & \xmark \\
Kontinuous Kontext~\cite{parihar2026kontinuous}
& \xmark & \cmark & \xmark \\
VeloEdit~\cite{li2026veloedit}
& \xmark & \cmark & \xmark \\
\midrule
\klein{} editor~\cite{bfl2026flux2klein}
& \cmark & \xmark & \xmark \\
\midrule
Ours
& \cmark & \cmark & \cmark \\
\bottomrule
\end{tabular}
\end{adjustbox}
\caption{Supported ways to specify and explore visual variation.
Multimodal specification accepts both text and visual measurements.
Attribute parameterizations map controls to attribute
values rather than edit strengths.
\cmark: supported; \xmark: absent from the compared formulation;
\pmark: available by scaling the fitted LoRA.}
\label{tab:intro_capability}
\end{table}

%% file: sec/2_related.tex
\section{Related Work}
\label{sec:related}

\input{fig/method-parameterizations}

\paragraph{Image editing.}
Text-to-image models can edit images from text instructions~\cite{brooks2023instructpix2pix} or visual references~\cite{bfl2026flux2klein}, or by noising and denoising generations~\cite{meng2022sdedit,hertz2023prompt,parmar2023zero}. Spatial adapters and guidance accept cues such as edges, depth or drags~\cite{zhang2023controlnet,epstein2023self,shi2024dragdiffusion}. Other methods infer an edit from before-and-after examples~\cite{nguyen2023visual,gu2024analogist,lu2025pairedit} or learn to edit parts from masks or examples~\cite{cvejic2025partedit,liu2025partcomposer}. These interfaces specify desired edits. Our focus is constructing reusable controls attached to prompt words whose settings expose
a parameterized range of continuous attribute values.

\paragraph{Continuous control.}
Text-space methods provide reusable, composable directions through token embeddings~\cite{baumann2025continuous}, contrastive prompt pairs~\cite{ekin2026unreasonable} or sparse autoencoder features~\cite{kamenetsky2025saedit}. Continuous controls have also been found in GAN latent spaces~\cite{harkonen2020ganspace,shen2020interfacegan,patashnik2021styleclip}, diffusion noise and $h$-space~\cite{kwon2023diffusion,park2023understanding,haas2024discovering,dalva2024noiseclr}, and LoRAs that decompose the generator's variation~\cite{gandikota2025sliderspace}. Concept Sliders~\cite{gandikota2024concept} learn sliders from text or paired visual examples; Text Slider~\cite{chiu2026textslider} and All-in-One Slider~\cite{ye2026allinone} learn controls from language. Kontinuous Kontext~\cite{parihar2026kontinuous}, SliderEdit~\cite{zarei2026slideredit,basu2026slidereditpp} and VeloEdit~\cite{li2026veloedit} expose edit strength in instruction-based editing. Concurrent work on UniSlider~\cite{serrano2026unislider} uses perceptual distances to source and target endpoints to learn intermediate edits and calibrate their progression. These methods generally parameterize edit strength or
perceptual progress toward an endpoint. Wizmo parameters
specify requested attribute values over domains suited
to the attribute, including cyclic and multidimensional
domains.

\paragraph{Guidance and rewards from visual evaluations.}
Visual evaluations can guide or supervise generation without curated edit pairs. Classifier guidance and its extensions steer sampling through gradients of image-based objectives~\cite{dhariwal2021diffusion,bansal2023universal,yu2023freedom}, while DOODL~\cite{wallace2023doodl} optimizes the initial noise. Reward fine-tuning updates model weights to improve learned scores~\cite{xu2023imagereward,clark2024draft,prabhudesai2023alignprop}, including feedback from VLMs~\cite{wang2026diffusiondrf,kumari2026learning}. Chang~\etal~\cite{chang2025instructing} optimize text embeddings against attribute classifiers, and Dual-Process~\cite{luo2025dual} distills a VLM's assessment of text and visual cues into a LoRA. Concurrent work on Measured Sliders~\cite{chen2026measured} learns measurement-guided LoRAs and calibrates their strength to realized image changes. Rather than steer generation at test time or optimize
toward a single target or higher score, we use visual
evaluations during fitting to match requested attribute
values across settings, requiring no evaluator at use
time. Fitting signals can come from geometric and image
losses, learned predictors and VLMs, or comparisons to
transformed and model-generated references whose
construction need not be differentiable.

\paragraph{Text states in diffusion transformers.}
Text conditioning offers several places to intervene in generation. P+~\cite{voynov2023p+} learns separate embeddings for different layers, TexTailor~\cite{li2026textailor} adjusts text hidden states across blocks, and SHIFT~\cite{konovalova2026shift} steers them with directions derived from prompt pairs. FluxSpace~\cite{dalva2025fluxspace} edits through text-conditioned attention outputs. Studies of MMDiTs show that contextualized text states encode visual attributes left unspecified by the prompt~\cite{dahary2026contextual}, attention outputs reveal concept locations~\cite{helbling2025conceptattention}, padding tokens affect generation~\cite{toker2025padding}, and particular layers play an important role in editing~\cite{avrahami2025stableflow}.
We build on these
observations by attaching wizmos to contextualized
prompt-word states. A learned mapping from attribute
parameters to offsets accommodates nonlinear visual
responses while retaining an explicit attachment to
the selected word.

%% file: fig/method-parameterizations.tex
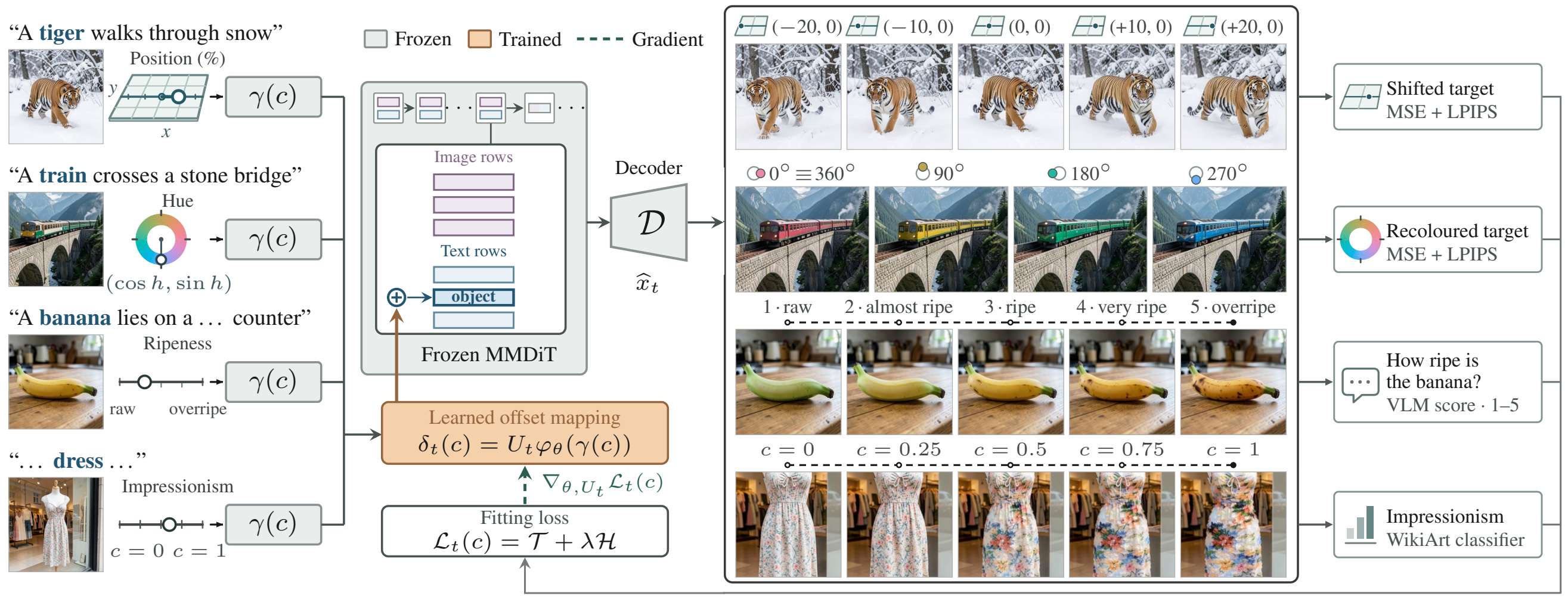
\begin{figure*}[t]
  \centering
  \resizebox{\textwidth}{!}{\input{fig/method-parameterizations-drawing}}
\caption{Constructing wizmos from visual evaluations.
Each row illustrates a separately fitted wizmo with a domain
and encoding suited to its attribute.
The learned offset mapping converts a parameter value $c$ into an update
to the selected word's contextualized text state in the frozen MMDiT.
Visual evaluations of the decoded prediction $\widehat{x}_t$
define the target loss $\mathcal T$, with optional preservation
loss $\mathcal H$.}
  \label{fig:method}
\end{figure*}

%% file: fig/method-parameterizations-drawing.tex
\begingroup%
\definecolor{mfInk}{HTML}{171819}%
\definecolor{mfMuted}{HTML}{48534D}%
\definecolor{mfFrozen}{HTML}{E9EEEC}%
\definecolor{mfFrozenLine}{HTML}{88928D}%
\definecolor{mfTrained}{HTML}{F5D1AB}%
\definecolor{mfTrainedLine}{HTML}{94653F}%
\definecolor{mfImage}{HTML}{EEE8F0}%
\definecolor{mfImageLine}{HTML}{816585}%
\definecolor{mfText}{HTML}{EAF2F7}%
\definecolor{mfTextLine}{HTML}{618AA2}%
\definecolor{mfSelected}{HTML}{D7E7F2}%
\definecolor{mfBlue}{HTML}{215572}%
\definecolor{mfGradient}{HTML}{28604E}%
\newcommand{\mfigmath}[1]{\setbox0=\hbox{$#1$}\dimen0=\dimexpr(\dp0-\ht0)/2\relax\raise\dimen0\box0}%
\newcommand{\mfigfont}[1]{\fontsize{#1bp}{8bp}\selectfont\rmfamily}%
\begin{tikzpicture}[x=1bp,y=-1bp,
  every node/.style={inner sep=0pt,outer sep=0pt,text=mfInk},
  mfline/.style={draw=mfMuted,line width=.65bp},
  mfarrow/.style={mfline,->,>=stealth},
  mfgrad/.style={draw=mfGradient,line width=.9bp,dash pattern=on 2.2bp off 1.8bp}]
\path[use as bounding box] (0,0) rectangle (484.4,187.4);
\begin{scope}[overlay]
\node[anchor=north west] at (0,0) {\includegraphics[width=484.4bp]{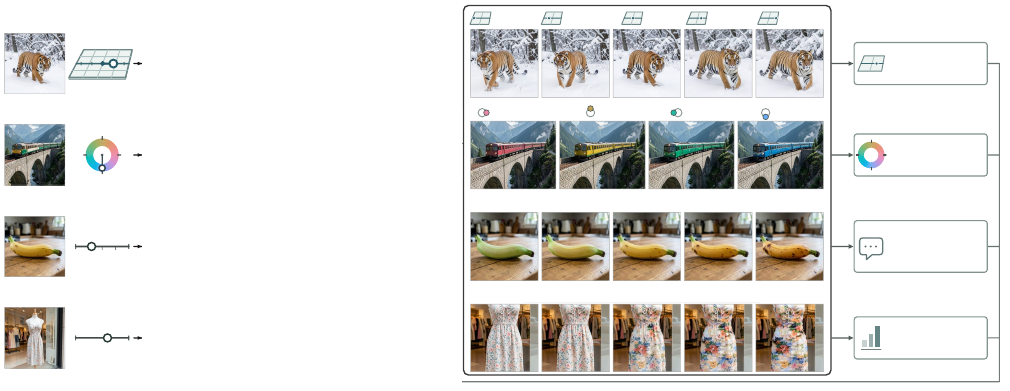}};

\foreach \cy in {30.54,74.44,118.34,162.24}{
  \draw[draw=mfFrozenLine,fill=mfFrozen,line width=.65bp,rounded corners=1.1bp]
    (69,\cy-6.1) rectangle (98.3,\cy+6.1);
  \draw[mfline] (98.3,\cy) -- (105.3,\cy);
}
\draw[mfline] (105.3,30.54) -- (105.3,162.24);
\draw[mfarrow] (105.3,134.5) -- (117,134.5);

\draw[draw=mfFrozenLine,fill=mfFrozen,line width=.6bp,rounded corners=.4bp] (111.7,10.1) rectangle (118.5,15.3);
\node[anchor=west] at (121,12.7) {\mfigfont{6.3}Frozen};
\draw[draw=mfTrainedLine,fill=mfTrained,line width=.6bp,rounded corners=.4bp] (143.7,10.1) rectangle (150.5,15.3);
\node[anchor=west] at (153,12.7) {\mfigfont{6.3}Trained};
\draw[mfgrad] (177,12.7) -- (191.3,12.7);
\node[anchor=west] at (194,12.7) {\mfigfont{6.3}Gradient};
\draw[draw=mfFrozenLine,fill=mfFrozen,line width=.65bp,rounded corners=2bp] (110.7,26) rectangle (180,116);

\foreach \xx in {114.5,127.5,146}{
  \draw[draw=mfFrozenLine,fill=white,line width=.4bp,rounded corners=.4bp] (\xx,29.4) rectangle (\xx+9,38.6);
  \draw[draw=mfImageLine,fill=mfImage,line width=.4bp] (\xx+1.2,30.8) rectangle (\xx+7.8,33.5);
  \draw[draw=mfTextLine,fill=mfText,line width=.4bp] (\xx+1.2,34.8) rectangle (\xx+7.8,37.3);
}
\draw[mfarrow,line width=.45bp] (123.8,34) -- (127.1,34);
\node at (141.25,34) {\mfigfont{6.4}$\cdots$};
\draw[mfarrow,line width=.45bp] (155.3,34) -- (160.6,34);
\draw[draw=mfFrozenLine,fill=white,line width=.4bp,rounded corners=.4bp] (161,29.4) rectangle (170,38.6);
\path[fill=mfImage] (162.2,32.6) rectangle (165.5,35.4);
\path[fill=mfText] (165.5,32.6) rectangle (168.8,35.4);
\draw[draw=mfFrozenLine,line width=.4bp] (162.2,32.6) rectangle (168.8,35.4);
\node at (176,34) {\mfigfont{6.4}$\cdots$};
\draw[mfline,line width=.5bp] (150.5,38.7) -- (150.5,45.2);

\draw[draw=mfMuted,fill=white,line width=.65bp,rounded corners=2bp] (115.2,45.2) rectangle (173.5,103.5);
\foreach \yy in {54.4,61.4,68.4}{
  \draw[draw=mfImageLine,fill=mfImage,line width=.55bp] (133,\yy) rectangle (157.5,\yy+4.8);
}
\foreach \yy in {82.9,89.9,96.9}{
  \draw[draw=mfTextLine,fill=mfText,line width=.55bp] (133,\yy) rectangle (157.5,\yy+4.8);
}
\draw[draw=mfBlue,fill=mfSelected,line width=.8bp] (133,89.9) rectangle (157.5,94.7);
\node[text=mfBlue] at (145.25,92.3) {\mfigfont{5.2}\bfseries object};
\node[text=mfImageLine] at (145.25,49.5) {\mfigfont{5.2}Image rows};
\node[text=mfBlue] at (145.25,78) {\mfigfont{5.2}Text rows};
\node at (150,109.7) {\mfigfont{6.5}Frozen MMDiT};

\draw[mfarrow] (180,69.2) -- (187.3,69.2);
\draw[draw=mfFrozenLine,fill=mfFrozen,line width=.75bp] (187.5,61.7) -- (210.7,57.5) -- (210.7,80.9) -- (187.5,76.7) -- cycle;
\node at (200,69.2) {\mfigfont{10.5}$\mathcal D$};
\node at (199.1,52.2) {\mfigfont{6.1}Decoder};
\draw[mfarrow] (211.1,69.2) -- (222.1,69.2);
\node at (199,87.8) {\mfigfont{7}$\widehat{x}_t$};

\draw[draw=mfTrainedLine,fill=mfTrained,line width=.65bp,rounded corners=2bp] (117,124.5) rectangle (205,143.5);
\node[text=mfTrainedLine] at (161,129.7) {\mfigfont{6.2}Learned offset mapping};
\node at (161,137.8) {\mfigfont{7.3}$\delta_t(c)=U_t\varphi_\theta(\gamma(c))$};
\draw[draw=mfTrainedLine,line width=.95bp,->,>=stealth] (121.5,124.5) -- (121.5,95.6);
\draw[draw=mfBlue,fill=white,line width=.75bp] (121.5,92.3) circle[radius=2.6bp];
\draw[draw=mfBlue,line width=.65bp] (119.9,92.3) -- (123.1,92.3);
\draw[draw=mfBlue,line width=.65bp] (121.5,90.7) -- (121.5,93.9);
\draw[draw=mfBlue,line width=.5bp,->,>=stealth] (124.8,92.3) -- (132.5,92.3);

\draw[draw=mfMuted,fill=white,line width=.65bp,rounded corners=2bp] (117,156.5) rectangle (205,172);
\node[text=mfMuted] at (161,160.6) {\mfigfont{6.1}Fitting loss};
\node at (161,167.7) {\mfigfont{7}$\mathcal L_t(c)=\mathcal T+\lambda\mathcal H$};
\draw[mfgrad,->,>=stealth] (161,155.9) -- (161,144.1);
\node[anchor=west,text=mfGradient] at (166,150) {\mfigfont{6}$\nabla_{\theta,U_t}\mathcal L_t(c)$};
\draw[draw={rgb,255:red,116;green,123;blue,118},line width=.55bp,->,>=stealth] (222.5,183.3) -- (161,183.3) -- (161,172.3);

\draw[draw=mfInk,line width=.57bp,dash pattern=on 2.05bp off 1.6bp]
  (241.8,100.0) -- (379.0,100.0);
\draw[draw=mfInk,line width=.57bp,dash pattern=on 2.05bp off 1.6bp]
  (241.8,143.9) -- (379.0,143.9);
\foreach \xx in {241.8,276.1,310.4,344.7}{
  \draw[draw=mfInk,fill=white,line width=.47bp] (\xx,100.0) circle[radius=.7bp];
  \draw[draw=mfInk,fill=white,line width=.47bp] (\xx,143.9) circle[radius=.7bp];
}
\draw[draw=mfInk,fill=mfInk,line width=.47bp] (379.0,100.0) circle[radius=.7bp];
\draw[draw=mfInk,fill=mfInk,line width=.47bp] (379.0,143.9) circle[radius=.7bp];
\foreach \xx/\txt in {241.8/{1\,\textperiodcentered\,raw},
  276.1/{2\,\textperiodcentered\,almost ripe},
  310.4/{3\,\textperiodcentered\,ripe},
  344.7/{4\,\textperiodcentered\,very ripe},
  379.0/{5\,\textperiodcentered\,overripe}}{
  \node[anchor=base] at (\xx,97.0) {\mfigfont{6}\color[HTML]{414743}\txt};
}
\foreach \xx/\txt in {241.8/{$c=0$},276.1/{$c=0.25$},310.4/{$c=0.5$},
  344.7/{$c=0.75$},379.0/{$c=1$}}{
  \node[anchor=base] at (\xx,140.9) {\mfigfont{6}\color[HTML]{414743}\txt};
}
\foreach \xx in {42.6,49.0,55.4}{
  \draw[draw=mfMuted,line width=.45bp] (\xx,160.8) -- (\xx,163.5);
}
\node[anchor=base west] at (2.2030,13.0050) {\fontsize{7bp}{8.4bp}\selectfont\rmfamily\color[HTML]{171819}\mdseries ``A \textcolor[HTML]{215572}{{\bfseries tiger}} walks through snow''};
\node[anchor=base] at (50.6020,43.0000) {\fontsize{6bp}{7.2bp}\selectfont\rmfamily\color[HTML]{425D62}\mdseries \textit{x}};
\node[anchor=base] at (34.2530,29.3000) {\fontsize{6bp}{7.2bp}\selectfont\rmfamily\color[HTML]{425D62}\mdseries \textit{y}};
\node[anchor=base] at (83.6520,30.4050) {\fontsize{7.7bp}{9.24bp}\selectfont\rmfamily\color[HTML]{171819}\mfigmath{\gamma(c)}};
\node[anchor=base west] at (237.0300,10.7700) {\fontsize{6bp}{7.2bp}\selectfont\rmfamily\color[HTML]{414743}\mdseries (\ensuremath{-}20, 0)};
\node[anchor=base west] at (271.2600,10.7700) {\fontsize{6bp}{7.2bp}\selectfont\rmfamily\color[HTML]{414743}\mdseries (\ensuremath{-}10, 0)};
\node[anchor=base west] at (309.8679,10.7700) {\fontsize{6bp}{7.2bp}\selectfont\rmfamily\color[HTML]{414743}\mdseries (0, 0)};
\node[anchor=base west] at (340.9060,10.7700) {\fontsize{6bp}{7.2bp}\selectfont\rmfamily\color[HTML]{414743}\mdseries (+10, 0)};
\node[anchor=base west] at (375.1360,10.7700) {\fontsize{6bp}{7.2bp}\selectfont\rmfamily\color[HTML]{414743}\mdseries (+20, 0)};
\node[anchor=base] at (54.2029,20.5050) {\fontsize{5.85bp}{7.02bp}\selectfont\rmfamily\color[HTML]{3B423E}\mdseries Position (\%)};
\node[anchor=base west] at (2.2029,56.9060) {\fontsize{7bp}{8.4bp}\selectfont\rmfamily\color[HTML]{171819}\mdseries ``A \textcolor[HTML]{215572}{{\bfseries train}} crosses a stone bridge''};
\node[anchor=base] at (83.6520,74.3050) {\fontsize{7.7bp}{9.24bp}\selectfont\rmfamily\color[HTML]{171819}\mfigmath{\gamma(c)}};
\node[anchor=base] at (51.5000,89.7000) {\fontsize{6bp}{7.2bp}\selectfont\rmfamily\color[HTML]{3C535C}\mdseries \ensuremath{(\cos h,\sin h)}};
\node[anchor=base west] at (236.2316,56.1110) {\fontsize{6bp}{7.2bp}\selectfont\rmfamily\color[HTML]{414743}\mdseries 0\ensuremath{^{\circ}}\hspace{.55bp}\ensuremath{\equiv}\hspace{.55bp}360\ensuremath{^{\circ}}};
\node[anchor=base west] at (286.9022,56.1110) {\fontsize{6bp}{7.2bp}\selectfont\rmfamily\color[HTML]{414743}\mdseries 90\ensuremath{^{\circ}}};
\node[anchor=base west] at (328.8272,56.1110) {\fontsize{6bp}{7.2bp}\selectfont\rmfamily\color[HTML]{414743}\mdseries 180\ensuremath{^{\circ}}};
\node[anchor=base west] at (370.9872,56.1110) {\fontsize{6bp}{7.2bp}\selectfont\rmfamily\color[HTML]{414743}\mdseries 270\ensuremath{^{\circ}}};
\node[anchor=base] at (54.2019,64.4050) {\fontsize{5.85bp}{7.02bp}\selectfont\rmfamily\color[HTML]{3B423E}\mdseries Hue};
\node[anchor=base west] at (2.2029,100.8050) {\fontsize{7bp}{8.4bp}\selectfont\rmfamily\color[HTML]{171819}\mdseries ``A \textcolor[HTML]{215572}{{\bfseries banana}} lies on a \ldots{} counter''};
\node[anchor=base] at (83.6519,118.2050) {\fontsize{7.7bp}{9.24bp}\selectfont\rmfamily\color[HTML]{171819}\mfigmath{\gamma(c)}};
\node[anchor=base west] at (33.2529,128.2550) {\fontsize{5.4bp}{6.5bp}\selectfont\rmfamily\color[HTML]{424B46}\mdseries raw};
\node[anchor=base east] at (69.3529,128.2550) {\fontsize{5.4bp}{6.5bp}\selectfont\rmfamily\color[HTML]{424B46}\mdseries overripe};
\node[anchor=base] at (54.2029,108.3050) {\fontsize{5.85bp}{7.02bp}\selectfont\rmfamily\color[HTML]{3B423E}\mdseries Ripeness};
\node[anchor=base west] at (2.2029,144.7050) {\fontsize{7bp}{8.4bp}\selectfont\rmfamily\color[HTML]{171819}\mdseries ``\ldots{} \textcolor[HTML]{215572}{{\bfseries dress}} \ldots{}''};
\node[anchor=base] at (83.6519,162.1050) {\fontsize{7.7bp}{9.24bp}\selectfont\rmfamily\color[HTML]{171819}\mfigmath{\gamma(c)}};
\node[anchor=base west] at (33.2528,172.1550) {\fontsize{6bp}{7.2bp}\selectfont\rmfamily\color[HTML]{424B46}\mdseries $c=0$};
\node[anchor=base east] at (69.3528,172.1550) {\fontsize{6bp}{7.2bp}\selectfont\rmfamily\color[HTML]{424B46}\mdseries $c=1$};
\node[anchor=base] at (54.2018,152.2050) {\fontsize{5.85bp}{7.02bp}\selectfont\rmfamily\color[HTML]{3B423E}\mdseries Impressionism};
\node[anchor=base west] at (426.1000,29.8550) {\fontsize{6.15bp}{7.38bp}\selectfont\rmfamily\mdseries\color[HTML]{181918}Shifted target};
\node[anchor=base west] at (426.1000,37.0550) {\fontsize{6bp}{7.2bp}\selectfont\rmfamily\mdseries\color[HTML]{48534D}MSE + LPIPS};
\node[anchor=base west] at (426.1000,73.7550) {\fontsize{6.15bp}{7.38bp}\selectfont\rmfamily\mdseries\color[HTML]{181918}Recoloured target};
\node[anchor=base west] at (426.1000,80.9550) {\fontsize{6bp}{7.2bp}\selectfont\rmfamily\mdseries\color[HTML]{48534D}MSE + LPIPS};
\node[anchor=base west] at (426.1000,113.6550) {\fontsize{6.15bp}{7.38bp}\selectfont\rmfamily\mdseries\color[HTML]{181918}How ripe is};
\node[anchor=base west] at (426.1000,120.2050) {\fontsize{6.15bp}{7.38bp}\selectfont\rmfamily\mdseries\color[HTML]{181918}the banana?};
\node[anchor=base west] at (426.1000,127.4050) {\fontsize{6bp}{7.2bp}\selectfont\rmfamily\mdseries\color[HTML]{48534D}VLM score \textperiodcentered{} 1--5};
\node[anchor=base west] at (426.1000,161.5550) {\fontsize{6.15bp}{7.38bp}\selectfont\rmfamily\mdseries\color[HTML]{181918}Impressionism};
\node[anchor=base west] at (426.1000,168.7550) {\fontsize{6bp}{7.2bp}\selectfont\rmfamily\mdseries\color[HTML]{48534D}WikiArt classifier};
\end{scope}
\end{tikzpicture}%
\endgroup%

%% file: sec/3_method.tex
\section{Method}
\label{sec:method}

\input{tab/sliders_define}

We realize each wizmo as a learned mapping from attribute
parameters to offsets in a selected prompt word's
contextualized text states. To fit the mapping, we
generate images across prompts, seeds and parameter
values, and backpropagate visual evaluation losses
through the frozen generator (\cref{fig:method}).
Once fitted, the mapping can be attached to other
applicable words and combined with other wizmos
without using the evaluator.

\subsection{Attribute control through prompt-word offsets}
\label{sec:method_rows}
\label{sec:method_network}

\paragraph{Attribute domain and encoding.}
A wizmo exposes a continuous attribute domain $\mathcal C$
for a selected prompt word. Users explore this domain by
choosing a parameter value $c\in\mathcal C$, which may be scalar or vector-valued. We encode this value as
$\gamma(c)$ so that the offset network's inputs reflect
the structure of the domain exposed to the user.

For example, a position wizmo exposes a two-dimensional
spatial region. Its parameter value $c=(c_x,c_y)$ contains two
coordinates, which we retain directly through $\gamma(c)=c$.
A hue wizmo exposes a color wheel, with one angular
parameter represented by a normalized scalar $c\in[0,1)$.
We encode it as
$\gamma(c)=(\cos(2\pi c),\sin(2\pi c))$,
giving the network a continuous, periodic representation
of the wheel.

\paragraph{Offset network.}
The offset network maps the encoded value $\gamma(c)$ to
an offset $\delta_t(c)$ for each selected denoising step $t$.
We choose these steps according to the attribute being
controlled~\cite{meng2022sdedit}.

Each wizmo has a mapping $\varphi_\theta$, with learned
weights $\theta$, shared across its selected denoising
steps, and a learned linear output layer $U_t$ for each
step:
\begin{equation}
    {\delta}_t(c)
    = U_t\,\varphi_\theta(\gamma(c)).
    \label{eq:path}
\end{equation}
The shared mapping combines a multilayer perceptron (MLP)
with two hidden layers and a learned linear combination
of fixed coordinate features, and both are learnt jointly.

\paragraph{Applying the offsets.}
We apply these offsets in the text stream of a multimodal
diffusion transformer (MMDiT), whose dual-stream blocks update
text and image states through joint
attention~\cite{esser2024sd3}. This lets contexualized text states acquire visual information about what they
describe~\cite{dahary2026contextual}.

In \klein{}, we modify the text states entering its
last dual-stream block. For a selected token of the target
word, let $h_t\in\mathbb R^d$ denote its hidden state
entering this block at denoising step $t$, where $d$ is
the state dimension. We call this state the token's
\emph{text row}.
We scale the offset by this state's
root-mean-square magnitude to obtain the updated row
$h'_t$:
\begin{equation}
    h'_t
    = h_t
    + \frac{\|h_t\|_2}{\sqrt d}
      \,\delta_t(c).
    \label{eq:offset}
\end{equation}
Attention propagates this change through the frozen
generator. \Cref{sec:supp_specify,sec:supp_fit} detail encodings, network
specifications, step schedules and
offset constraints.

\subsection{Constructing wizmos from visual evaluations}
\label{sec:method_fit}
\label{sec:method_eval}

To construct a wizmo, we define its attribute domain
$\mathcal C$ and the meaning of its parameter values,
choose an encoding $\gamma(c)$ suited to the domain,
and select a visual evaluation that measures agreement
with the target attribute values. We fit the same
offset mapping across generated examples with different
objects, prompts and seeds, using a small set of generated source images (\cref{sec:eval}), to encourage it to generalize
beyond the fitting images.

\input{fig/applications}

\paragraph{Defining the fitting signal.}
We define a target loss $\mathcal T(x,c)$ that measures
how well an image $x$ exhibits the attribute value
specified by $c$. For example, hue can be specified by an absolute angle, while size can be specified relative to the unedited source image.

When a differentiable evaluator $E(x)$ measures the
attribute, $\mathcal T$ penalizes the discrepancy between
its measurement and the target value. We apply this loss
at intermediate parameter values as well as endpoints.

Sometimes, it is easier to find or generate references
depicting desired attribute values than to measure the
attribute directly. In this case, we associate references with parameter values and define $\mathcal T$ through weighted pixel
differences and perceptual distances to the corresponding
reference. For progression toward a fixed reference,
$c$ instead specifies the target comparison loss.
References can be transformed source images, selected
generations from an initial fit, or outputs of stronger
editing models, distilling their attribute-specific
guidance into a wizmo.
Further, reference construction and value assignment need not be
differentiable; the comparison supplies the gradients.
\Cref{tab:sliders_define} summarizes the target losses
used to fit our wizmos.

\paragraph{Fitting the mapping.}
The active denoising steps are fixed for each attribute.
For each training prompt, seed and parameter value $c$, at each selected step $t$, we apply $\delta_t(c)$ to the
selected prompt-word state using \cref{eq:offset} and decode
the step's clean-image prediction,
$\widehat{x}_t(\delta_t(c))$.

An optional preservation loss $\mathcal H(x)$, weighted by $\lambda$, compares
pixels or features with the corresponding unmodified
prediction at the same step.
For one training example at denoising step $t$,
we minimize
\begin{equation}
    \mathcal L_t(c)
    =
    \mathcal T\bigl(\widehat{x}_t(\delta_t(c)),c\bigr)
    +
    \lambda\mathcal H\bigl(\widehat{x}_t(\delta_t(c))\bigr).
    \label{eq:loss}
\end{equation}
We backpropagate each loss through the evaluation,
decoder and current generator step, holding its incoming
latent fixed. Gradients do not propagate between denoising
steps; we accumulate them before updating $\theta$
and $\{U_t\}$. The generator and evaluation models
remain fixed. \Cref{sec:appendix_impl} provides
attribute-specific losses, reference construction,
sampling and optimization details.

\subsection{Using wizmos}
\label{sec:method_units}

\paragraph{Applying a wizmo.}
A fitted wizmo stores its domain $\mathcal C$, active
denoising steps and network weights $(\theta,\{U_t\})$,
with no fixed prompt or token index. Users attach it to an applicable prompt word and select a parameter value $c\in\mathcal C$.
Wizmos act within the generator's usual denoising process,
applying $\delta_t(c)$ to the selected word's text rows
at the active steps (\cref{eq:offset}).
This adds only evaluation of the small offset network
and scaling and addition of its outputs; no extra
generator passes, evaluators or optimization are required
(\cref{sec:appendix_cost}).

To explore a trajectory $c(\tau)\in\mathcal C$, such as
a hue arc or a curved spatial route, users sample values
along it and render each with the prompt
and seed held fixed (\cref{fig:sliders}).
Here, $\tau$ indexes exploration, not denoising.
Intermediate parameter
values require no new references or refitting.

\input{fig/composition}

\input{fig/sliders}

\paragraph{Combining wizmos.}
Multiple wizmos compose in one denoising run,
without joint fitting or sequential re-editing.
Those attached to different words update separate text rows.
On a shared row, let $k$ index the wizmos active at step $t$.
Their offsets combine as
$\delta_t=\sum_k w_k\,\delta_{k,t}(c_k)$ in \cref{eq:offset}.
Unit weights ($w_k=1$) compose attributes such as size and
style. Varying the weights creates exploration spaces
(\cref{fig:applications}): a scalar weight controls offset
strength, while barycentric weights
($w_k\ge0$, $\sum_k w_k=1$) blend wizmos across a simplex.
The parameters $c_k$ specify attribute values,
while the weights $w_k$ scale or blend offsets.

%% file: tab/sliders_define.tex
\begin{table}[t]
\centering
\footnotesize
\setlength{\tabcolsep}{4pt}
\begin{adjustbox}{max width=\linewidth}
\begin{tabular}{@{}lll@{}}
\toprule
Wizmo & Attribute scale & Fitting signal \\
\midrule
\tgroup{3}{Geometry}
Size & relative size & transformed references \\
Position & image-width fraction & transformed references \\
Horizon height & image-height fraction & transformed references \\
Sketch silhouette & mask overlap (IoU) & sketch comparison (SAM~3) \\
\midrule
\tgroup{3}{Appearance}
Hue & hue angle ($^\circ$) & recoloured references \\
Palette & CIELAB distance & palette comparison (OT) \\
Cartoon style & reference progress & restyled references \\
Impressionism & style probability & WikiArt classifier \\
\midrule
\tgroup{3}{Perceptual and semantic}
Fill level & fullness level & generated references \\
Ripeness & graded level & VLM scores (Gemma~4) \\
Weathering & graded level & VLM scores (Gemma~4) \\
Saliency & fixation share & DeepGaze IIE predictor \\
\bottomrule
\end{tabular}
\end{adjustbox}
\caption{Wizmos grouped by attribute type.
Attribute scale gives the reported units;
fitting signal identifies the references or evaluations
used for construction. Graded levels use ordered verbal descriptions as anchors;
VLM probabilities yield a continuous expected level. Reference progress measures
reduction in the reference-comparison loss.
OT denotes optimal transport.}
\label{tab:sliders_define}
\end{table}

%% file: fig/applications.tex
\begin{figure}
\centering
\includegraphics[width=\columnwidth]{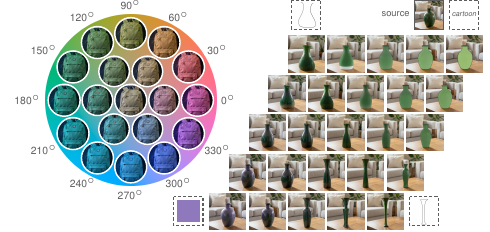}
\caption{Exploration spaces assembled from fitted wizmos.
Angle specifies the backpack's hue, while radius
scales the offset. Sketch, hue and style wizmos combine for a vase design space.}
\label{fig:applications}
\label{fig:exploration_space}
\label{fig:exploration}
\end{figure}

%% file: fig/composition.tex
\begin{figure}[t]
\centering
\includegraphics[width=\columnwidth]{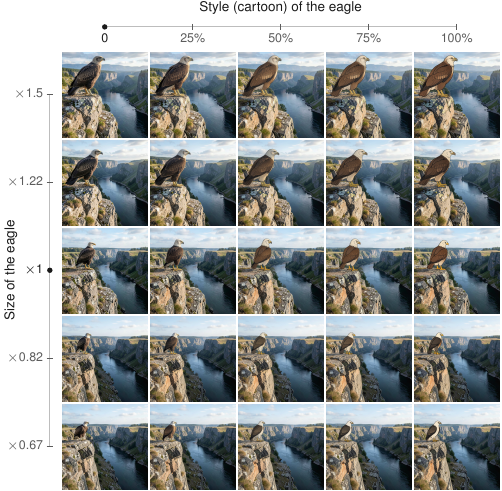}
\caption{Size and style wizmos, separately fitted
on cats, vases and cars, are applied together to ``eagle''
without refitting.}
\label{fig:composition}
\end{figure}

%% file: fig/sliders.tex
\begin{figure*}[t]
\centering
\includegraphics[width=\textwidth]{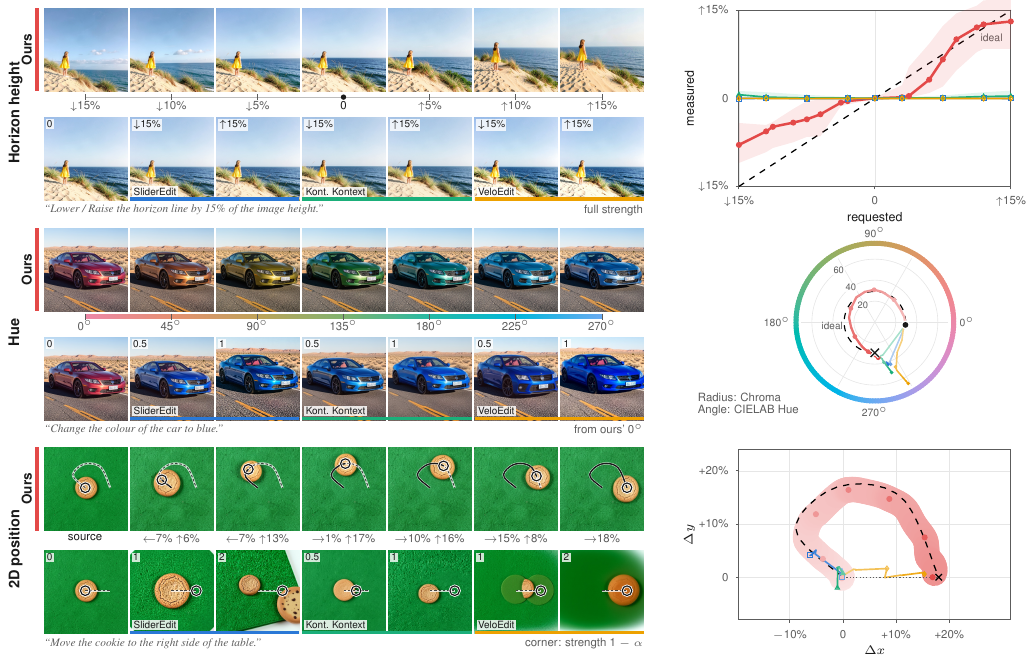}
\caption{Scalar, cyclic and multidimensional attribute
parameterizations. Wizmos explore horizon height,
a $270^\circ$ hue arc and a curved two-dimensional
position path. Plots visualize
attribute adherence. Baselines receive instructions
shown beneath the images.}
\label{fig:sliders}
\end{figure*}

%% file: sec/4_experiments.tex
\input{fig/gallery}

\section{Results}
\label{sec:eval}

Object attributes in product and industrial design are
often classified across geometry, appearance and
semantics~\cite{herriott2021aesthetics}.
We thus demonstrate wizmos across this spectrum
(\cref{tab:sliders_define} and \cref{fig:gallery}), including parameterizations unsupported by prior art, such as cyclic hue and
two-dimensional position controls (\cref{fig:sliders}).
On tasks supported by all methods, wizmos more accurately realize
requested attribute values (\cref{tab:comparison}).
Fitted wizmos transfer beyond their fitting objects:
a fill-level wizmo fitted on liquids can control the amount of LEGOs in a jar (\cref{fig:gallery}). Reuse extends to new prompts and
objects in real images, hue and cartoon-style wizmos edit a photographed bird and lamp, without refitting
(\cref{fig:gallery}).
Separately fitted wizmos also compose on and across
objects without joint fitting (\cref{fig:composition,fig:exploration}).
We further demonstrate applicability across generators
with wizmos separately fitted on FLUX.3~[klein],
shown in \cref{sec:supp_second_generator}.

\paragraph{Setup.}
Our main experiments use \klein{}~\cite{bfl2026flux2klein}
4B at $768^2$ with four denoising steps.
Each wizmo uses 9--81 source generations, typically 27.
Size and Palette use three objects and Hue nine,
each with three prompts and three seeds per prompt.
Reference-based fitting uses transformed or edited
versions of these sources; no manually curated
source--target training pairs are required.
Wizmos require no additional generator passes.
On one H100, five wizmos on five objects add negligible
rendering overhead with unchanged peak memory.
We evaluate our one-dimensional controls on eleven-point
sweeps, including values not sampled during fitting.
All qualitative results shown use unseen prompts,
including new object words, without refitting (\cref{fig:gallery,fig:sliders,fig:composition}).
\Cref{sec:appendix_cost,sec:supp_measurement} detail costs and evaluation
protocols.

\paragraph{Baselines.}
No baseline covers our full range of specifications
and parameterizations.
We compare Concept Sliders~\cite{gandikota2024concept},
Dual-Process~\cite{luo2025dual},
SliderEdit~\cite{zarei2026slideredit},
Kontinuous Kontext~\cite{parihar2026kontinuous},
VeloEdit~\cite{li2026veloedit} and the \klein{} reference
editor on size, hue and palette, which all can attempt.  For palette control for instance, the editors receive instructions naming the palette's colours, Concept Sliders
uses those colours in prompt pairs, and Dual-Process's fitting VLM receives palette swatches. Text-space steering~\cite{baumann2025continuous,
ekin2026unreasonable,kamenetsky2025saedit} additionally
requires converting instructions into descriptive prompt
contrasts; these adapted comparisons appear in
\cref{sec:supp_comparisons} (\cref{tab:comparison_text}).
Comparisons use common sources and measurements,
retaining each method's backbone, fitting procedure
and input format.
For sweeps, each method's maximum comparison strength
represents the full target change; intermediate strengths
are scored against corresponding attribute values.

\paragraph{Metrics.}
Parameter error (PE) measures deviation from target
attribute values at the endpoints; PE$_{\mathrm{sweep}}$
averages this error over the sweep, in the attribute's
reported units. We use independent evaluators for attribute error unless no independent measurement is available on the reported
scale; such as an absolute comparison, like hue, in which case we reuse the function. We therefore assess the intended edit also through independent visual question answering (VQA) for edit adherence (VQA$_{\mathrm{edit}}$) and identity
preservation (VQA$_{\mathrm{id}}$)~\cite{serrano2026unislider},
and monotonic progression (MP) judgments of attribute
ordering~\cite{basu2026slidereditpp}, all from a VLM
not used for fitting.
Following UniSlider~\cite{serrano2026unislider}, the
coefficient of variation (CV) of consecutive
DreamSim~\cite{fu2023dreamsim} distances measures
perceptual evenness.
Background change ($\Delta$BG) compares features outside
the edited object, following HATIE~\cite{ryu2025hatie},
using DINOv3~\cite{simeoni2025dinov3}.
We assess image quality with
MUSIQ~\cite{ke2021musiq,serrano2026unislider},
reporting its change from the source.
\Cref{sec:appendix_impl,sec:supp_measurement} provide all fitting signals and evaluators.

\subsection{Attribute coverage and parameterizations}
\label{sec:eval_range}
\label{sec:eval_quality}
\label{sec:eval_trajectories}

We show qualitative comparisons in \cref{fig:sliders}, with plots on the right comparing measured attribute
changes against the targets,
and additional results of our method in \cref{fig:gallery}.

A common fitting procedure constructs wizmos from
measurements, predictors and visual references
(\cref{tab:sliders_define}).
For example, silhouette control uses a differentiable
comparison between SAM~3's soft mask and a target sketch
to define progress toward the sketch (\cref{fig:gallery}).
For hue, source images recoloured to sampled CIELAB
angles provide targets for a differentiable
image-comparison loss (\cref{fig:sliders}).
Ripeness uses five verbal anchors: raw, almost ripe,
ripe, very ripe and overripe (\cref{fig:method}).
The VLM's probabilities over these anchors yield a
continuous expected level, allowing parameter
values to target levels between the anchors.

Horizon height, hue and two-dimensional position
illustrate scalar, cyclic and multidimensional attribute
domains, respectively (\cref{fig:sliders}).
Instruction-based baselines expose textual edits and
strength parameters, without calibration to attribute coordinates or direct control over intermediate values and paths. Even instructions to shift the horizon by $15\%$ of
image height leave it largely unchanged.
Wizmos learn this calibration through visual evaluations:
our hue wizmo follows a $270^\circ$ arc through colours
skipped by the baselines' shorter routes to blue, while
our position wizmo follows a curve instead of their
roughly straight route to the endpoint.

\subsection{Accuracy, progression and preservation}
\label{sec:eval_comp}

Quantitative results are summarized in \cref{tab:comparison}.
Wizmos achieve the best endpoint and sweep accuracy on
the compared tasks. Their sweep results show closer
agreement with target attribute values across the
evaluated range, supporting exploration beyond endpoints.

The \klein{} reference editor achieves the next-highest
edit adherence and the best identity and background
preservation (\cref{tab:comparison}), but does not sweep over attribute parameters.
Wizmos improve attribute accuracy and edit adherence
over this with only modest preservation costs,
while providing reusable continuous control over attributes.

\input{tab/comparison}

\subsection{Transfer and composition}
\label{sec:eval_properties}

Fitted wizmos retain strong edit adherence on unseen
prompts without refitting (\cref{tab:comparison}).
Transfer includes new object words: for e.g, size, fitted on cats, vases and cars, transfers to
the globe without refitting (\cref{fig:gallery}); and realizing a semantic concept meaningfully
across objects: a fill-level wizmo fitted on liquids
expresses fullness by adding or removing LEGO bricks
in a jar. Reuse extends to real photographs after
inversion, where hue and cartoon-style wizmos edit
a bird and lamp without refitting (\cref{fig:gallery}). Tests in \cref{sec:supp_transfer} further show comparable quantitative performance on new unseen objects.
SliderEdit achieves better background and identity
preservation on transfer, potentially because it makes
less of the intended change, as indicated by its lower
edit-adherence score and the largely unchanged horizon
in \cref{fig:sliders}.
Separately fitted wizmos compose on the unseen eagle
(\cref{fig:composition}) and support exploration
(\cref{fig:exploration}), with less cross-object leakage
than sequential instruction-based edits
(\cref{sec:supp_compose}).

\subsection{Ablation studies}
\label{sec:eval_ablation}

\paragraph{Offset mapping.}
We compare our learned mapping with fixed
learned offsets scaled by $c$ ($c\,u$).
In the middle image column ($c=0.5$) of
\cref{fig:ablation}, ours produces an intermediate
palette, while $c\,u$ leaves the blouse nearly unchanged.
The graph plots palette change against $c$, normalized
by each method's change at $c=1$.
Our mapping follows the dashed diagonal's proportional
progression more closely, whereas the scaled offset
changes abruptly.

\paragraph{Intervention representation.}
A low-rank adapter (LoRA) provides another compact
representation for learning an edit under the same visual
supervision. It places the intervention in model
weights rather than a selected word's text state.
We scale these updates by the parameter value and compare
with the scaled-offset variant, holding linear scaling
fixed. In the garment example, word offsets target either
garment, while the tested LoRA changes both
(\cref{fig:ablation}).
For Ripeness, it reaches the VLM's target by redrawing
the scene, while offsets edit the banana in place.

\input{fig/ablation}

%% file: fig/gallery.tex
\begin{figure*}[t]
\centering
\includegraphics[width=\textwidth]{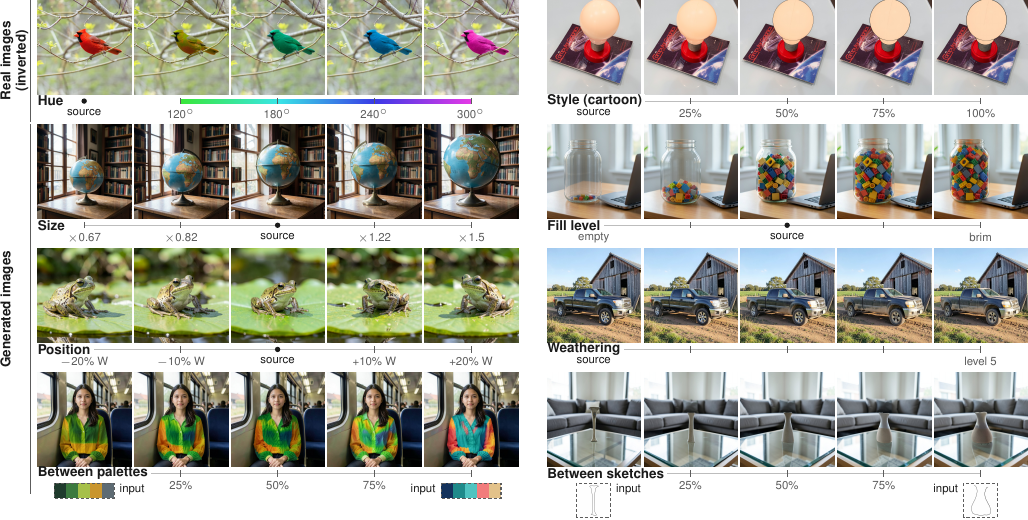}
\caption{Wizmos across the attribute spectrum.
Fitted wizmos vary geometry, appearance and semantic
properties on real photographs (top row) and on new generations. Labels show size
multipliers, position relative to frame width $W$,
fill levels and weathering levels. Percentages mark
progression toward a target appearance or between
the displayed palette and sketch references.}
\label{fig:gallery}
\end{figure*}

%% file: tab/comparison.tex
\begin{table*}[t]
\centering
\footnotesize
\setlength{\tabcolsep}{3.4pt}
\begin{adjustbox}{max width=\linewidth}
\begin{tabular}{@{}l ccc c cc ccc c c c ccc ccc}
\toprule
 & \multicolumn{4}{c}{Endpoint} & \multicolumn{5}{c}{Sweep} & Local & Identity & Quality & \multicolumn{6}{c}{Transfer} \\
\cmidrule(lr){2-5}\cmidrule(lr){6-10}\cmidrule(l){14-19}
 & \multicolumn{3}{c}{PE\down} & & & & \multicolumn{3}{c}{PE$_\text{sweep}$\down} & & & & \multicolumn{3}{c}{PE$_\text{sweep}$\down} & & & \\
\cmidrule(lr){2-4}\cmidrule(lr){8-10}\cmidrule(lr){14-16}
Method & Size & Hue & Palette & VQA$_\text{edit}$\up & MP\up & CV\down & Size & Hue & Palette & $\Delta$BG\down & VQA$_\text{id}$\up & $\Delta$MUSIQ\up & Size & Hue & Palette & VQA$_\text{edit}$\up & VQA$_\text{id}$\up & $\Delta$BG\down \\
\midrule
\tgroup{19}{Fitting-based sliders}
Concept Sliders~\cite{gandikota2024concept} & \sworst 0.91 & 17.8 & 19.0 & 0.76 & \sworst 0.65 & 0.79 & \sworst 0.55 & 36.3 & \second 9.6 & \worst 0.075 & \worst 0.69 & \sworst $-$3.8 & \sworst 0.58 & 42.6 & \second 9.6 & 0.75 & \worst 0.68 & \sworst 0.064 \\
Dual-Process~\cite{luo2025dual} & \worst 1.02 & \worst 108.6 & \sworst 24.7 & \worst 0.42 & \worst 0.51 & \best 0.57 & \worst 0.61 & \worst 59.3 & \sworst 13.7 & \sworst 0.074 & \sworst 0.79 & \best 0.1 & \worst 0.62 & \worst 61.8 & \sworst 12.7 & \worst 0.46 & \sworst 0.79 & \worst 0.066 \\
\midrule
\tgroup{19}{Instruction-based sliders}
SliderEdit~\cite{zarei2026slideredit} & 0.68 & 11.3 & 22.8 & 0.74 & \sworst 0.65 & \worst 1.63 & \second 0.45 & \sworst 49.0 & \worst 14.6 & \second 0.026 & 0.89 & \second $-$0.2 & 0.47 & 50.3 & \worst 14.7 & 0.74 & \second 0.89 & \second 0.027 \\
Kontinuous Kontext~\cite{parihar2026kontinuous} & \second 0.61 & 11.0 & \worst 27.5 & \sworst 0.71 & 0.66 & \sworst 1.12 & \second 0.45 & 47.8 & 11.0 & 0.043 & 0.86 & $-$3.3 & 0.46 & \sworst 53.9 & 10.4 & \sworst 0.70 & 0.85 & 0.049 \\
VeloEdit~\cite{li2026veloedit} & 0.69 & \second 10.1 & 20.6 & 0.74 & \best 0.75 & 0.86 & \second 0.45 & \second 32.6 & 10.4 & 0.035 & 0.86 & \worst $-$4.0 & \second 0.42 & \second 39.3 & \second 9.6 & 0.74 & 0.85 & 0.036 \\
\midrule
\tgroup{19}{Reference}
\klein editor~\cite{bfl2026flux2klein} & 0.89 & \sworst 19.0 & \second 14.6 & \second 0.83 & \na & \na & \na & \na & \na & \best 0.010 & \best 0.92 & \second $-$0.2 & \na & \na & \na & \second 0.80 & \best 0.90 & \best 0.008 \\
\midrule
\tgroup{19}{Ours}
Ours & \best 0.29 & \best 9.6 & \best 11.4 & \best 0.87 & \second 0.72 & \second 0.72 & \best 0.22 & \best 11.4 & \best 7.4 & 0.037 & \second 0.90 & $-$0.3 & \best 0.30 & \best 17.8 & \best 8.9 & \best 0.82 & 0.85 & 0.034 \\
\bottomrule
\end{tabular}
\end{adjustbox}
\caption{Accuracy, progression and preservation on shared
Size, Hue and Palette tasks. PE and PE$_{\text{sweep}}$
measure endpoint and sweep errors in normalized log scale,
degrees and CIELAB distance, respectively. Other scores
average the three tasks. Transfer uses new prompts without
refitting learned controls. Green and red mark the best
and worst results; dark and light shades indicate first
and second.}
\label{tab:comparison}
\end{table*}

%% file: fig/ablation.tex
\begin{figure}[t]
\centering
\includegraphics[width=\columnwidth]{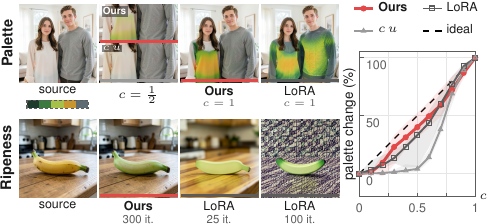}
\caption{Learned mappings improve palette tracking over fixed
offsets. Word offsets target the blouse; LoRA changes both
garments. Under identical VLM ripeness supervision (bottom),
offsets edit in place; LoRA redraws the scene by 25 iterations.}
\label{fig:ablation}
\end{figure}

%% file: sec/5_conclusion.tex
\section{Conclusion}

We introduced wizmos: reusable controls attached to prompt
words for continuous exploration of visual attributes.
They tie exploration to meaningful attribute values and
paths, while supporting transfer across generations and
composition on and across objects.

Their attainable quality and range of edits, however, remain limited
by the generator's prior. Continuous visual change need
not imply a physically meaningful progression: weathering
levels (\cref{fig:gallery}), for example, need not correspond
to elapsed time or realistic material aging. Future work
could explore physically grounded temporal progressions
and functional edits such as opening a drawer.

In their current form, wizmos let users build their own
control interfaces by distilling guidance from measurements,
references, predictors or larger generative models into
controls for their own generators.
As generators and visual priors become more capable, this abstraction offers
a way to navigate their expanding visual diversity and understanding through
meaningful, reusable controls.

%% file: sec/A_appendix.tex
\clearpage
\appendix
\setcounter{topnumber}{2}
\setcounter{dbltopnumber}{2}
\renewcommand{\topfraction}{0.99}
\renewcommand{\dbltopfraction}{0.99}
\renewcommand{\textfraction}{0.01}
\renewcommand{\floatpagefraction}{0.7}
\renewcommand{\dblfloatpagefraction}{0.7}
\twocolumn[
  \centering
  \Large
  \textbf{\thetitle}\\
  \vspace{0.5em}Appendix\\
  \vspace{1.0em}
]
\raggedbottom
This appendix describes how we constructed the wizmos in the paper
and what we learned from fitting them across a range of visual attributes.
For each control, we define what its parameter values mean, how to
measure the corresponding visual change, and how to fit its word
offsets (\cref{sec:appendix_impl}).

We examine how the fitted controls track their parameter values,
transfer to new objects and compose with other edits
(\cref{sec:appendix_eval}). We then compare methods on shared editing
tasks, including the additional text-space steering comparison
promised in the main paper (\cref{sec:supp_comparisons}). Ablations
examine the fitting choices that affect accuracy and locality,
alongside fitting and rendering costs and new fits on a second
generator (\cref{sec:supp_fitting_analysis}).

The project page\footnote{\url{https://www.cs.toronto.edu/~zhaoqi/wizmos/}} includes an interactive gallery spanning the
wizmos and editing results. Readers can vary parameter values,
compare objects and methods, and explore combinations of controls
(\cref{sec:supp_examples}). Evaluation sources, sweeps and metric
definitions are collected in \cref{sec:supp_measurement}.

\medskip
\begingroup
\setlength{\fboxsep}{6pt}
\noindent\colorbox{cvprblue!6}{\parbox{\dimexpr\columnwidth-2\fboxsep\relax}{%
\small\textbf{For a visual overview.}
Start with reference construction (\cref{fig:ref_construction},
p.\,\pageref{fig:ref_construction}), object transfer
(\cref{fig:generality}, p.\,\pageref{fig:generality}), or method
comparisons (\cref{fig:baselines},
pp.\,\pageref{fig:baselines}--\pageref{fig:baselines_continued}).
For interactive sweeps and combinations, see the project page
(\cref{sec:supp_examples}).}}
\endgroup
\medskip

\section{Constructing wizmos}
\label{sec:appendix_impl}

Constructing a wizmo requires a parameter domain, a fitting target for
each parameter value, and a mapping from these values to word offsets.
We describe the domains and attribute-specific losses first, then how
we optimize the mapping across prompts, seeds and denoising steps.
\Cref{fig:ref_construction} illustrates the reference construction;
\cref{tab:steps,tab:fitting} collect the fitting configurations.

\subsection{Attribute scales and parameter domains}
\label{sec:supp_specify}

A parameter value $c\in\mathcal C$ specifies an absolute attribute
value, a change relative to the source, or progress toward a visual
target. Source-relative scalar controls use $c=0$ for the source,
with $[0,1]$ for progress toward one endpoint or $[-1,1]$ for changes
in two directions. The units and endpoint meanings are defined with
each fitting objective in \cref{sec:supp_losses}.

\begin{figure}[t]
\centering
\includegraphics[width=\columnwidth]{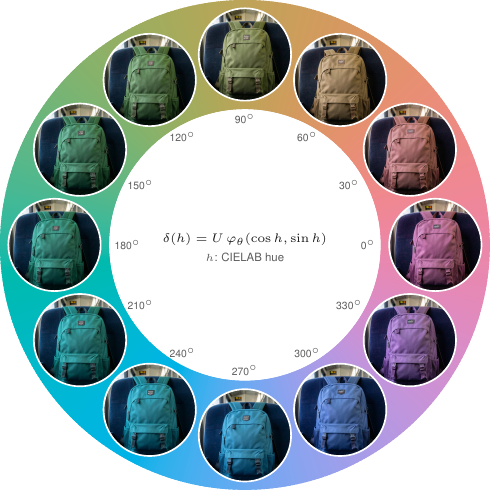}
\caption{Exploring the full colour wheel.
One fitted wizmo renders a backpack at twelve requested CIELAB hue angles.
Encoding $h$ as $(\cos h,\sin h)$ closes the path, giving the same
offset at $0^\circ$ and $360^\circ$.
The outer ring marks the requested hues.}
\label{fig:parameterization}
\end{figure}

The encoding reflects the domain. Hue uses $c\in[0,1)$ for an
absolute CIELAB angle $360^\circ c$, independent of the source hue.
Writing $h=2\pi c$, its encoding $(\cos h,\sin h)$ returns to the
same offset after a full turn (\cref{fig:parameterization}). Joint Position passes
a coordinate pair $(c_x,c_y)$ to one fitted mapping, which can follow
a two-dimensional path without refitting (\cref{fig:sliders}). The
quantitative Position benchmark evaluates horizontal displacement.

\subsection{Fitting targets and losses}
\label{sec:supp_losses}

The fitting objective connects the parameter scale to a measurable
visual change. Depending on the attribute, it compares the prediction
with a reference image, measures progress toward a visual target, or
matches a predictor score (\cref{tab:sliders_define}). We evaluate it on
the decoded clean-image prediction $\widehat x_t$ at an active
denoising step $t$. Its unedited counterpart $\widehat x_t^{\mathrm{src}}$
comes from the same prompt and seed with all offsets disabled.
For source-relative score targets, we apply the same evaluator to
$\widehat x_t^{\mathrm{src}}$ to obtain the baseline at that step.
This makes zero requested change agree with the unedited prediction,
even when its score varies over denoising.

\begin{figure*}[t]
\centering
\includegraphics[width=\textwidth]{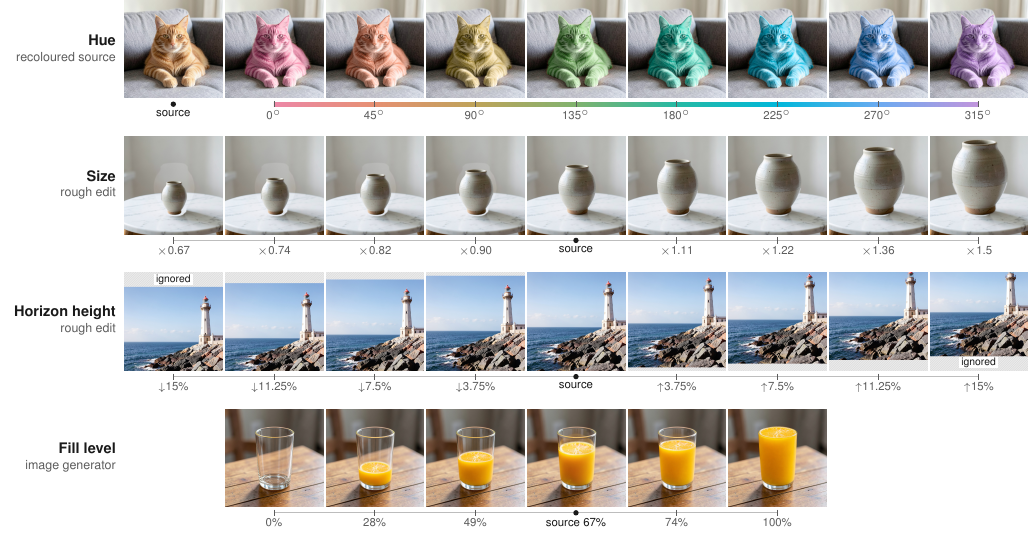}
\caption{Constructing references.
Each row shows a fitting source and its references at several parameter values, made by recolouring (Hue), rough
edits (Size, Horizon height) or an image generator (Fill level).
The horizon's loss ignores the band that the shift leaves empty.}
\label{fig:ref_construction}
\end{figure*}

\input{tab/steps}

\paragraph{References at specified parameter values.}
Image references are constructed from the final unedited source
generation. Size and Position use a SAM~3 object cutout, composited
over a background from which \klein's editor has removed the object.
Size scales the cutout by $1.5^c$, so $c=-1$ and $c=1$ request
factors $1/1.5$ and $1.5$. Horizontal Position translates it by
$0.2c$ of the image width, with positive values moving right.
Horizon height shifts the frame downward by $0.15c$ of its height,
excluding the uncovered band from the loss. These three controls use
$c\in[-1,1]$ and nonzero fitting references at quarter increments.

Joint Position uses $(c_x,c_y)\in[-1,1]^2$, with displacements of
$0.2c_x$ times the image width and $0.2c_y$ times the image height;
positive values move right and upward, respectively. Its references
use axis-aligned and diagonal displacements at one-third, two-thirds
and full range.

For Hue, we recolour the masked object to CIELAB angle $360^\circ c$,
preserving lightness and near-achromatic pixels. Reference hues are
rounded to $5^\circ$ increments; recoloured pixels have chroma at
least 20 where the sRGB gamut permits. These transformations provide
targets on the requested geometry or colour scale
(\cref{fig:ref_construction}).

For Fill level, we normalize measured fill fractions to $c\in[-1,1]$.
We generate five edits per vessel with GPT Image through Codex and
use Codex to estimate their fill fractions.
Let $f_0$ be the final source's measured fraction and $f_-,f_+$ those
of the generated empty and full references. An edit with fraction
$f$ receives parameter
\begin{equation*}
c=\begin{cases}
(f-f_0)/(f_+-f_0), & f\geq f_0,\\
(f-f_0)/(f_0-f_-), & f<f_0.
\end{cases}
\end{equation*}
Thus $c=0$ denotes the source and $c=-1,1$ the two reference
endpoints. We paste each edit inside the vessel and omit references
with $|c|<0.15$. Fitting uses these discrete references; the learned
mapping supplies intermediate parameter values.
\Cref{fig:ref_construction} shows the reference construction.

For these controls, the reference for a given $c$ is shared across
active steps. Size, Position, Fill level and Hue use the image
comparison
\begin{equation*}
    L(\widehat x_t,y)
    = 10\,\frac{\sum_p w_p\,\|\widehat x_t(p)-y(p)\|_2^2}{3\sum_p w_p}
    + \mathrm{LPIPS}\bigl(\widehat x_t|_B,\,y|_B\bigr),
\end{equation*}
where $y$ is the reference, $B$ is the edited region's bounding box,
and pixels in that region receive five times the weight of other
pixels. For Horizon height, we compare the prediction's change from
its unedited counterpart with the reference's change from the source,
using the pixel term after Gaussian blurring (32 pixels).

\paragraph{Progress toward a visual target.}
For cartoon style, Palette and Sketch silhouette, $c\in[0,1]$
specifies progress in a comparison loss from the source at $c=0$
to the learned endpoint at $c=1$.
For cartoon style, \klein's editor restyles an object crop and we
paste it back inside the mask, using the image comparison above.
Palette uses entropic optimal transport from garment pixels, weighted
by SAM~3's soft mask, to equal masses on the five swatches in CIELAB
with $\Delta L^*$ halved. The mask weights are held fixed during
backpropagation through this loss.

For Sketch silhouette we align the target silhouette $A$ to the
object's box and compare it with the SAM~3 soft mask $p$. The loss
combines Dice overlap $D$ at three blur scales, row-width error,
and overlap in a band $b$ along the outline,
\begin{equation*}
\begin{aligned}
    L ={}& \sum_{\sigma\in\{0,8,24\}}\bigl(1-D(G_\sigma*p,\,G_\sigma*A)\bigr)\\
    &+ 3\,\frac{\sum_y\bigl|\sum_x s(p)-\sum_x A\bigr|}{\sum A}
    + 1-D(p\,b,\,A\,b),
\end{aligned}
\end{equation*}
where $s(p)=\operatorname{sigmoid}(12(p-\frac12))$.

For all three controls, let $L_{0,t}$ and $L_{1,t}$ denote the source
and endpoint comparison losses at step $t$. We minimize $L_{1,t}/L_{0,t}$
at the endpoint. For $0<c<1$, we use
\begin{equation*}
\left(\frac{L_t(c)-[(1-c)L_{0,t}+cL_{1,t}]}{L_{0,t}}\right)^2,
\end{equation*}
treating the current iteration's endpoint loss as fixed when taking
gradients. Normalizing by $L_{0,t}$ accounts for different source-loss
magnitudes across images and denoising steps.

\paragraph{Predictor scores.}
For Impressionism and Saliency, a predictor supplies the attribute
score used for fitting. A SigLIP~2 classifier fine-tuned on WikiArt
scores Impressionism on the isolated object. For $c\in[0,1]$, we
interpolate the target log-odds as
$\ell_t^\star=(1-c)\ell_{0,t}+c\,\operatorname{logit}(0.9)$,
where $\ell_{0,t}$ is the unedited prediction's log-odds.
For predicted log-odds $\ell$ the loss is $((\ell-\ell_t^\star)/s_t)^2$,
with $s_t=\max(\operatorname{logit}(0.9)-\ell_{0,t},1)$.

For Saliency, $c\in[-1,1]$ controls the share of DeepGaze~IIE's
predicted fixations inside the object. We minimize squared error
against endpoint targets $\pm0.15$ (15 percentage points) from the
source share at step $t$. For intermediate
parameter values, the target change is $|c|$ times the change reached
by the endpoint of the same sign in that iteration, averaged over
sources, steps and views. We clip target shares to $[0.02,0.98]$.

\begin{figure*}[t]
\centering
\includegraphics[width=\textwidth]{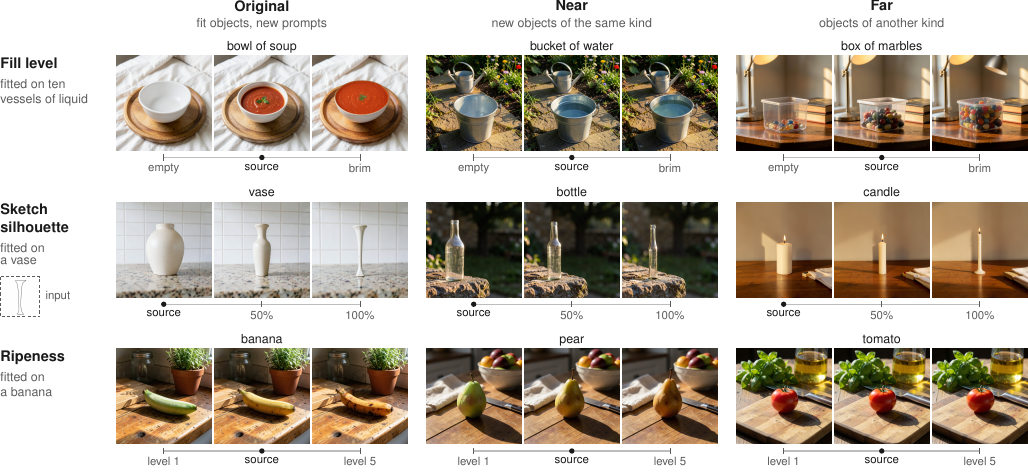}
\caption{Object transfer without refitting. Columns show the Original,
Near and Far tiers in \cref{tab:generality}. Fill level was fitted on
ten vessels (Table~\ref{tab:generality}\subref{tab:transfer_multi_category}).
Sketch silhouette was fitted on vases and Ripeness on bananas
(Table~\ref{tab:generality}\subref{tab:transfer_single_category}).
In the Fill level and Ripeness rows, the source lies between
the two requested endpoints. The Sketch silhouette row shows the
source and edits at $c=0.5$ and $c=1$ for the sketch at left.}
\label{fig:generality}
\end{figure*}

\paragraph{Ordered semantic levels.}
For Ripeness and Weathering, five ordered verbal descriptions define
the attribute scale. Gemma~4 (12B) gives a probability $p_k$ for each
description, and $\sum_k k\,p_k$ yields a continuous expected level.
We minimize squared error against a target $|c|$ of the way from
the step's source score toward the relevant endpoint.
Ripeness uses $c\in[-1,1]$: negative values progress toward raw
(level~1), and positive values toward overripe (level~5). Weathering
uses $c\in[0,1]$ and progresses toward level~5.
The Ripeness anchors are \emph{unripe and green} (raw in the main paper),
\emph{almost ripe}, \emph{ripe}, \emph{very ripe, with brown spots}
and \emph{overripe and brown}.
Weathering uses \emph{brand new}, \emph{slightly used}, \emph{used and
somewhat worn}, \emph{old and weathered} and \emph{very old, worn out
and decaying}.

\input{tab/fitting}

\input{tab/sliders}

\subsection{Fitting the word offsets}
\label{sec:supp_fit}

We fit one parameter-to-offset mapping per wizmo, shared across its
source prompts and seeds. The generator and evaluators remain frozen.
The following choices determine how that mapping acts on a word and
how it is optimized against the losses above.

\paragraph{Mapping and word binding.}
On \klein{} 4B, offsets enter the last dual-stream block at the text
rows associated with the selected prompt word. We locate its occurrence
by character span; for a word split into subwords, the offset is
broadcast over its rows or applied to its final row.
The mapping combines fixed coordinate features with an MLP of two
width-64 hidden layers and SiLU activations. It outputs $K$ coefficients
($K=4$ for one-dimensional domains and $K=8$ for joint Position), which a learned
$U_t\in\mathbb R^{3072\times K}$ maps to an offset at each active step
(\cref{eq:path}).

For source-relative controls, subtracting the raw offset at $c=0$
enforces a neutral origin. We clip the offset's RMS magnitude to
$\rho$, then scale it by the source word state's RMS magnitude at
that step. The source magnitudes are cached from the unedited
generation and remain fixed across the parameter sweep. Hue uses
its periodic encoding without origin subtraction.

\paragraph{Step-wise fitting.}
We generate $768^2$ images in four denoising steps. Early offsets can
change layout; starting at steps~2--4 lets the source layout form
before changing appearance. \Cref{tab:steps} lists the active steps
and reference or calibration used by each wizmo.
At each active step, we evaluate the loss on the decoded clean-image
prediction, holding the incoming latent fixed during differentiation.
The edited latent is then passed to the next step. Gradients from
the active steps, sampled parameter values and fitting sources are
accumulated for each update.

\paragraph{Sources, optimization and preservation.}
\Cref{tab:fitting} collects the fitting sources, iteration counts,
output-layer learning rates, offset bounds and preservation terms.
Adam uses $\beta=(0.9,0.999)$ and a shared-mapping learning rate of
$0.01$. Both learning rates decay on a cosine schedule, except for
Palette, which uses constant rates. Preservation terms compare the
region outside the object with the unedited prediction at the same
step. We retain the final iterate.

Joint Position uses 27 top-down sources spanning apples, mugs and
cookies. It is fitted for 200 iterations on axis-aligned references,
then 100 with diagonal references added, at the first denoising step
with $\rho=4$ and output-layer learning rate $0.12$.
\Cref{sec:appendix_ablation} examines the effect of source diversity;
\cref{sec:appendix_cost} reports fitting and rendering costs.

\paragraph{Applying fitted wizmos to photographs.}
Photographs can be used through latent inversion or the generator's
reference-image input. Inversion maps a photograph to a
noise initialization; reference-image conditioning
instead supplies the photograph as an input to the generator.
In either case, the fitted wizmo applies its offsets to the selected
prompt word without refitting. The prompt and initialization remain
fixed across each parameter sweep. The main-paper gallery shows
this reuse for Hue and cartoon style on photographs
(\cref{fig:gallery}).

\section{Properties of fitted wizmos}
\label{sec:appendix_eval}
\label{sec:supp_reuse}

We examine how a fitted wizmo tracks its parameter values, transfers
beyond its fitting objects, and interacts with other controls. These
analyses cover the range of wizmos constructed in
\cref{sec:appendix_impl}: attribute tracking in \cref{tab:sliders},
object transfer in \cref{fig:generality,tab:generality}, and composition
in \cref{tab:composition}.

\begingroup
\expandafter\let\expandafter\supplementTableStar\csname table*\endcsname
\expandafter\def\csname table*\endcsname[#1]{\supplementTableStar[!tp]}
\input{tab/generality}
\endgroup

\subsection{Attribute tracking}
\label{sec:appendix_results}

Across the full set of wizmos, \cref{tab:sliders} reports how closely
each control tracks its parameter values on its own attribute scale.
Each sweep varies the parameter while holding the prompt and seed
fixed. Test uses new seeds of fitting prompts; Transfer uses new
prompts and includes new object words. The common sweep grid and
sample counts are specified once in \cref{sec:appendix_protocol}.

Endpoint error (PE) measures how closely the edit reaches its target,
while sweep error (PE$_{\mathrm{sweep}}$) measures agreement along
the way. These errors retain the units of their attribute, so we
compare them within each row. The separate ordering measure asks
whether two frames move in the intended direction.
\Cref{sec:supp_metrics} defines the measures and identifies the
cases that reuse the fitting signal at evaluation.

Hue and Ripeness are ordered consistently on held-out seeds.
For Position and Horizon height, both VQA$_{\mathrm{edit}}$ and MP rely on locating the object centre or horizon relative to labelled reference lines. Small displacements can be difficult for the VLM to resolve, with ties receiving half credit in MP. We therefore interpret these scores alongside geometry-based parameter errors and the visible trajectories in \cref{fig:sliders}, with
further interactive sweeps on the project page
(\cref{sec:supp_examples}).
Saliency has modest VQA$_{\mathrm{edit}}$ and near-chance ordering scores. For Saliency, the predicted fixation-share target does not
prescribe a unique visual change: its sweeps under \emph{Every wizmo}
on the project page (\cref{sec:supp_examples}) show added object
detail and stronger colour contrast as the parameter increases.
These are plausible attention cues, but their relative salience can
be ambiguous to a VLM. The sweeps therefore help interpret the VLM
judgments alongside agreement with DeepGaze~IIE's prediction.

\input{tab/composition}

\subsection{Transfer across objects}
\label{sec:supp_transfer}

We test whether a fixed mapping retains the meaning of its attribute
when attached to new object words. Original uses fitting categories
in new scenes; Near and Far use related and more distant categories
for which the attribute remains meaningful. \Cref{tab:generality}
separates transfer after fitting on multiple categories from the
stronger generalization test of fitting on just one category.
The object sets are listed in the table, with sampling described in
\cref{sec:appendix_protocol}.

Across both panels, edit adherence decreases as the objects differ
more from those used for fitting. The single-category tests make
this especially visible for Ripeness: a control fitted on bananas
is less reliable on other fruits. Horizon height retains similar
endpoint error across its scene categories.
\Cref{fig:generality} illustrates both successful transfer and its
limits: Fill level changes the amount of solid contents but leaves
some at the empty endpoint, Sketch silhouette partly reproduces the
target outline, and the Far Ripeness endpoints look similar.

Transfer can also require a different visual expression of the same
attribute. In the main-paper gallery, Fill level changes the amount
of LEGO bricks in a jar after fitting on liquid-filled vessels
(\cref{fig:gallery}). The cartoon-style wizmo similarly carries its
treatment to an unseen eagle (\cref{fig:composition}). These examples
complement the category-wise analysis in \cref{tab:generality}.

\subsection{Composition on one or more objects}
\label{sec:supp_compose}

A scene can contain two edits on one object or a different edit on
each of several objects. On one word we sum the active offsets with
unit weights. Across words we add each offset to its own selected
word. Both arrangements take one denoising pass.
\Cref{tab:composition} tests two attributes on one object and one
attribute per object in two- and three-object scenes.

We assess both intended changes and leakage: whether an object
changes along an attribute assigned to another object. Wizmos retain
high intended-edit scores while producing less cross-object leakage
than the sequential editors in these tests (\cref{tab:composition}).
SliderEdit is evaluated in its sequential configuration.
\Cref{sec:supp_metrics} defines the object-level scores and application
protocols.

\section{Comparisons with other methods}
\label{sec:supp_comparisons}

\begin{figure*}[t]
\centering
\includegraphics[width=\textwidth,trim=0 223.428bp 0 0,clip]{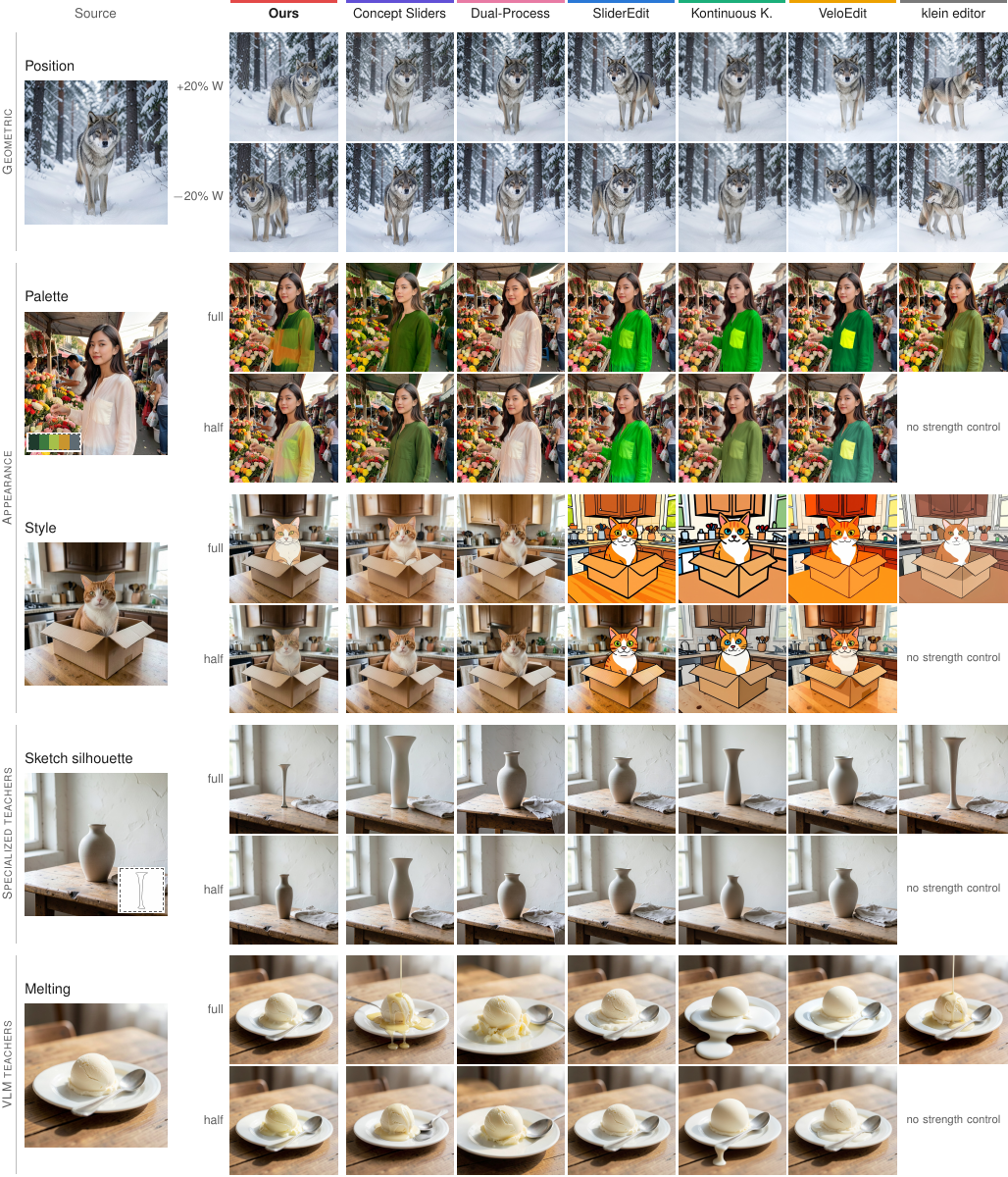}
\caption{Qualitative comparisons using each method's supported inputs.
Position shows opposite endpoints; Palette and cartoon style compare
full and half strength where supported. Instructions appear beneath
the examples. The reference editor supplies endpoints.
Sketch silhouette and melting continue on p.\,\pageref{fig:baselines_continued}.}
\label{fig:baselines}
\end{figure*}

\begin{figure*}[t]
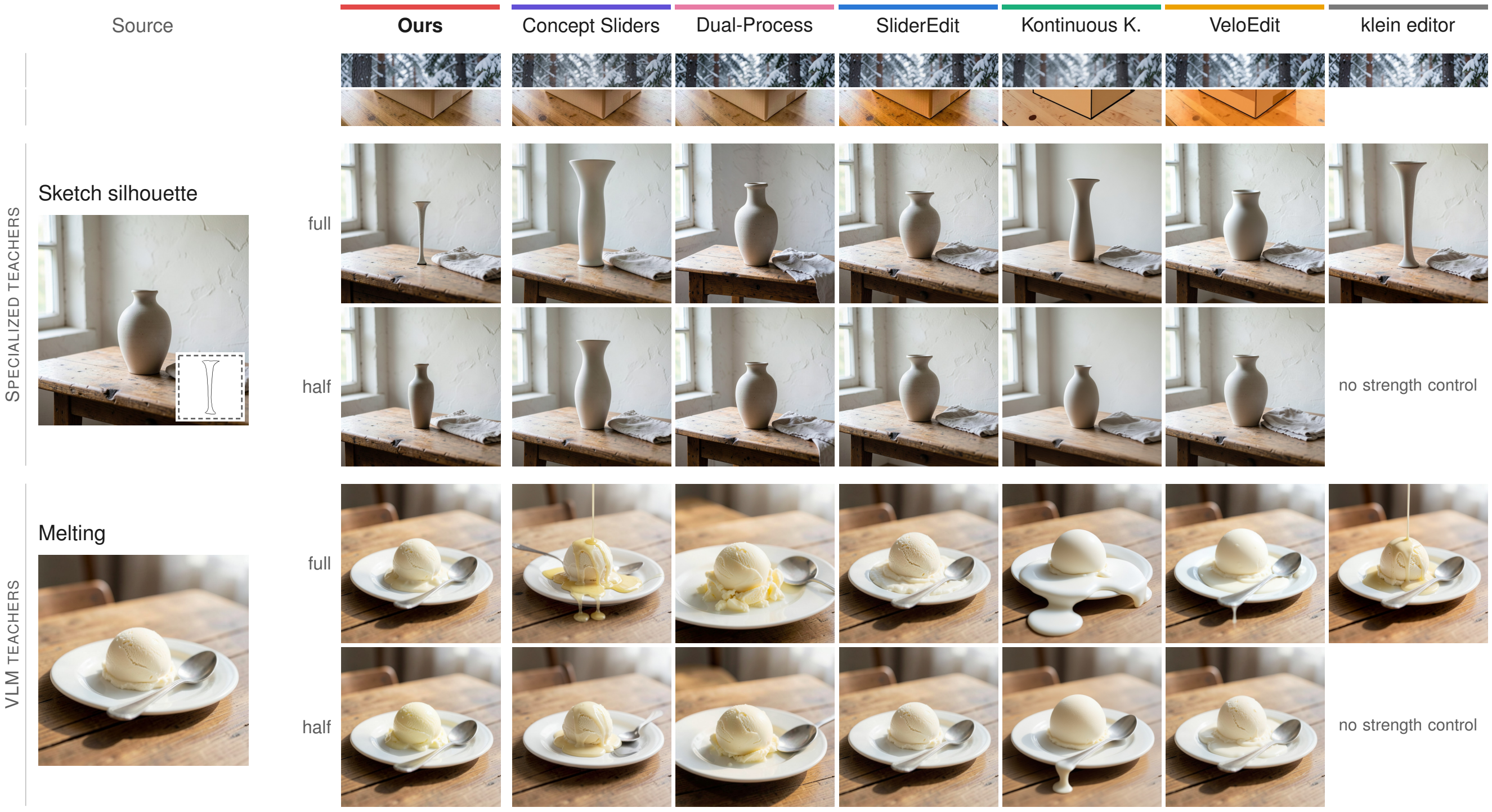

\ContinuedFloat
\centering
\includegraphics[width=\textwidth,trim=0 548.428bp 0 0,clip]{fig/pdf/baselines.pdf}\par
\includegraphics[width=\textwidth,trim=0 0 0 337bp,clip]{fig/pdf/baselines.pdf}
\caption{Qualitative comparisons (continued). Sketch silhouette and
melting at full and half strength where supported. Instructions
appear beneath the examples; the reference editor also receives the
sketch and supplies an endpoint. Melting is an additional qualitative
example outside the benchmark in \cref{tab:sliders}.}
\label{fig:baselines_continued}
\end{figure*}

\input{tab/comparison_text}

\Cref{fig:baselines} extends the visual comparisons in the main
paper, while \cref{tab:comparison_text} adds the promised text-space
steering comparison. Size, Hue and
Palette provide geometric and appearance edits that all compared
methods can attempt through their supported interfaces.

\paragraph{Common tasks and inputs.}
All methods are evaluated on the same sources against a target scale
factor, hue angle or palette. Wizmos and text-space steering use the
source prompt and seed; image editors receive its pixels and an edit
instruction or reference. For Palette, instruction-based editors
receive colour names, Concept Sliders uses names in prompt pairs,
and Dual-Process receives swatches through its fitting VLM.

We treat each method's maximum comparison strength as the full target
change and assess intermediate strengths against the corresponding
fraction. The visual comparisons show half and full strength:
1.25 and 2.5 for Concept Sliders, 0.5 and 1 for Dual-Process, and
the corresponding strengths for instruction-based controls. The
reference editor supplies an endpoint. Post-hoc calibration that
keeps these endpoints fixed can improve intermediate tracking but
does not change endpoint error or preservation; we report endpoint
and sweep measures separately.

\begin{figure*}[t]
\centering
\newcommand{\depthmarks}{%
  \begin{pgfonlayer}{axis background}
    \fill[gray!12] ({axis cs:-1,0}|-{rel axis cs:0,0}) rectangle ({axis cs:4.5,0}|-{rel axis cs:0,1});
  \end{pgfonlayer}
  \draw[black, dotted] ({axis cs:4,0}|-{rel axis cs:0,0}) -- ({axis cs:4,0}|-{rel axis cs:0,1});}
\newcommand{\streamnames}{%
  \node[gray, font=\tiny, anchor=south] at ({axis cs:1.75,0}|-{rel axis cs:0,0.02}) {dual-stream};
  \node[gray, font=\tiny, anchor=south] at ({axis cs:11.25,0}|-{rel axis cs:0,0.02}) {single-stream};}
\begin{tikzpicture}
\begin{groupplot}[
  group style={group size=2 by 3, horizontal sep=1.3cm, vertical sep=12pt,
    xticklabels at=edge bottom, xlabels at=edge bottom, ylabels at=edge left},
  set layers, width=0.48\textwidth, height=2.8cm,
  xmin=-1, xmax=18, xtick={0,2,4,9,17},
  tick label style={font=\scriptsize}, label style={font=\footnotesize}, title style={font=\footnotesize, yshift=-3pt},
  scaled y ticks=false, yticklabel style={/pgf/number format/fixed},
  grid=major, grid style={gray!20},
  every axis plot/.append style={thick, mark size=1.2pt, error bars/y dir=both, error bars/y explicit,
    error bars/error bar style={thin, solid}, error bars/error mark options={rotate=90, mark size=1.2pt, solid}},
]
\nextgroupplot[ymin=0.6, ymax=1.02, ytick={0.6,0.8,1}, ylabel={VQA$_\text{edit}$\up}, title={Size}]
\depthmarks\streamnames
\addplot[cvprblue, mark=*] coordinates {(0,0.842) += (0,0.058) -= (0,0.062) (2,0.870) += (0,0.035) -= (0,0.033) (4,0.898) += (0,0.031) -= (0,0.031) (9,0.854) += (0,0.046) -= (0,0.051) (17,0.716) += (0,0.046) -= (0,0.045)};
\nextgroupplot[ymin=0.6, ymax=1.02, ytick={0.6,0.8,1}, title={Hue}]
\depthmarks\streamnames
\addplot[orange!85!black, mark=*] coordinates {(0,0.855) += (0,0.114) -= (0,0.137) (2,0.843) += (0,0.111) -= (0,0.128) (4,0.900) += (0,0.077) -= (0,0.088) (9,0.986) += (0,0.013) -= (0,0.021) (17,0.998) += (0,0.001) -= (0,0.001)};
\nextgroupplot[ymin=0.4, ymax=1.02, ytick={0.4,0.6,0.8,1}, ylabel={VQA$_\text{id}$\up}]
\depthmarks
\addplot[cvprblue, mark=*] coordinates {(0,0.513) += (0,0.105) -= (0,0.092) (2,0.533) += (0,0.133) -= (0,0.127) (4,0.591) += (0,0.112) -= (0,0.119) (9,0.600) += (0,0.111) -= (0,0.105) (17,0.700) += (0,0.081) -= (0,0.097)};
\nextgroupplot[ymin=0.4, ymax=1.02, ytick={0.4,0.6,0.8,1}]
\depthmarks
\addplot[orange!85!black, mark=*] coordinates {(0,0.873) += (0,0.050) -= (0,0.053) (2,0.918) += (0,0.028) -= (0,0.028) (4,0.920) += (0,0.030) -= (0,0.033) (9,0.938) += (0,0.019) -= (0,0.020) (17,0.970) += (0,0.009) -= (0,0.011)};
\nextgroupplot[ymin=0.05, ymax=0.11, ytick={0.06,0.08,0.1}, ylabel={$\Delta$BG\down}, xlabel={Layer at which the offset is added}]
\depthmarks
\addplot[cvprblue, mark=*] coordinates {(0,0.083) += (0,0.013) -= (0,0.012) (2,0.087) += (0,0.014) -= (0,0.013) (4,0.081) += (0,0.012) -= (0,0.012) (9,0.070) += (0,0.012) -= (0,0.010) (17,0.069) += (0,0.012) -= (0,0.011)};
\nextgroupplot[ymin=0, ymax=0.03, ytick={0,0.01,0.02,0.03}, xlabel={Layer at which the offset is added}]
\depthmarks
\addplot[orange!85!black, mark=*] coordinates {(0,0.018) += (0,0.007) -= (0,0.005) (2,0.012) += (0,0.004) -= (0,0.003) (4,0.008) += (0,0.002) -= (0,0.002) (9,0.010) += (0,0.003) -= (0,0.002) (17,0.006) += (0,0.002) -= (0,0.001)};
\end{groupplot}
\end{tikzpicture}
\caption{Where to apply the offset, for Size (left) and Hue (right).
The layer is the input of \klein{}'s block at which the offset is added, counted from 0 over its five dual-stream blocks (shaded) and 20 single-stream blocks; layer 0 is the prompt embedding.
Size has its strongest mean edit response at layer 4, the last dual-stream block (dotted); Hue remains effective deeper in the network.
Points show transfer-set means with $95\%$ bootstrap intervals.}
\label{fig:ablation_where}
\end{figure*}
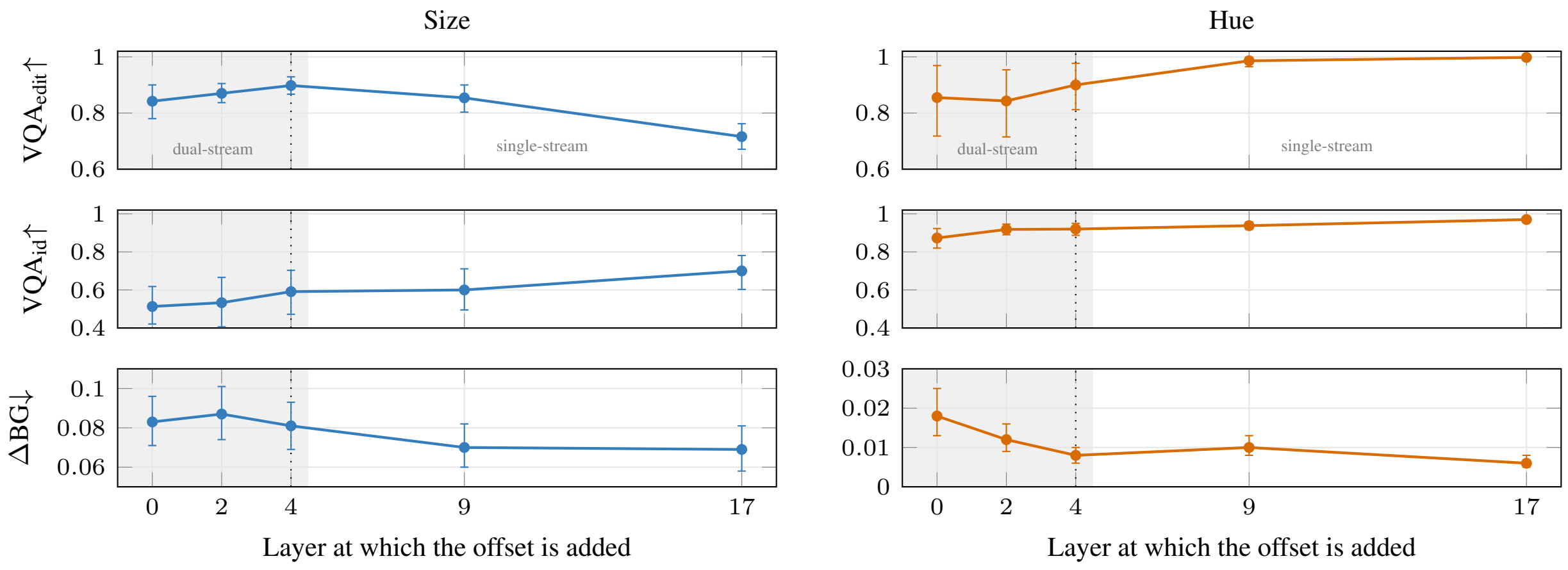

\paragraph{Text-space steering.}
Both additional methods derive an editing direction from contrasting
prompt descriptions. Ekin \& Gandelsman~\cite{ekin2026unreasonable}
use 100 prompt pairs and apply the difference of their mean encodings
to language-model-selected tokens. Their endpoint is the largest
tested strength within DreamSim distance 0.15 of the source; we test
default and extended searches reaching $8\times$ and $64\times$ the
starting strength. Baumann et al.~\cite{baumann2025continuous} learn
a direction on the object token from prompt triples built from our
fitting prompts. At their default strength of 2, we compare their
usual onset after one fifth of the steps with steering from the first
step.

Earlier steering improves Size and Hue endpoint accuracy for Baumann
et al., with reduced identity preservation. Extending the search has
little effect on Hue error for Ekin \& Gandelsman. Across these
variants, wizmos have lower endpoint and sweep errors on all three
attributes (\cref{tab:comparison_text}).

\paragraph{Related approaches.}
SAEdit~\cite{kamenetsky2025saedit} requires a sparse autoencoder for
the generator's text encoder; the available autoencoders did not
support \klein{}. Measured Sliders~\cite{chen2026measured} is
concurrent work whose code was not publicly available at the time
of our experiments, preventing an empirical comparison.
NumeriKontrol~\cite{xu2025numerikontrol} and
AttriCtrl~\cite{chen2026attrictrl} also study attribute-valued control.
Our focus is on word-attached controls fitted through visual
evaluations over scalar, cyclic and multidimensional attribute domains.

\section{Analysis of the fitting procedure}
\label{sec:supp_fitting_analysis}

\subsection{Ablations of fitting choices}
\label{sec:appendix_ablation}

The main paper illustrates the effects of learning the offset mapping
and attaching it to a word (\cref{fig:ablation}). Here we quantify
those effects and test how intervention depth and the variety of fitting
prompts and seeds affect the edit
(\cref{fig:ablation_where,tab:ablation,tab:ablation_sources}).
The same procedure is then fitted
on a second generator (\cref{sec:supp_second_generator}).

\input{tab/ablation}

\paragraph{Where to intervene.}
We use Size and Hue to test how intervention depth affects control
over object geometry and appearance. Size asks the generator to move
the object's boundary and fill any newly exposed background. Hue asks
it to change colour within that boundary while preserving shape and
position. For each wizmo we add the word offset at the input of
block 0, 2, 4, 9 or 17. The first three
are dual-stream blocks and the last two are single-stream blocks.
Size has its strongest mean edit response at the last dual-stream
block (4), while Hue also responds deeper in the network
(\cref{fig:ablation_where}). At the chosen depth,
\cref{tab:ablation} compares the target word with offsets added to
the image tokens inside the object mask or to every image token.
Masked image tokens yield the lowest Size endpoint error. Both
image-token variants have larger Palette errors than word offsets.

\input{tab/ablation_sources}

\paragraph{What to fit.}
Full versus scaled word offsets ($c\,u$) tests parameter dependence
while retaining word locality. The learned mapping improves endpoint
and sweep accuracy for Size and Palette (\cref{tab:ablation}).

We compare the scaled LoRA with the scaled word offset to hold the
linear dependence on $c$ fixed. LoRA has lower endpoint errors but
higher cross-object leakage (0.77 versus 0.14). This locality difference
is therefore present even without the nonlinear mapping.
The main-paper ablation shows Palette and Ripeness (\cref{fig:ablation}).

\begin{figure*}[t]
\centering
\includegraphics[width=\textwidth]{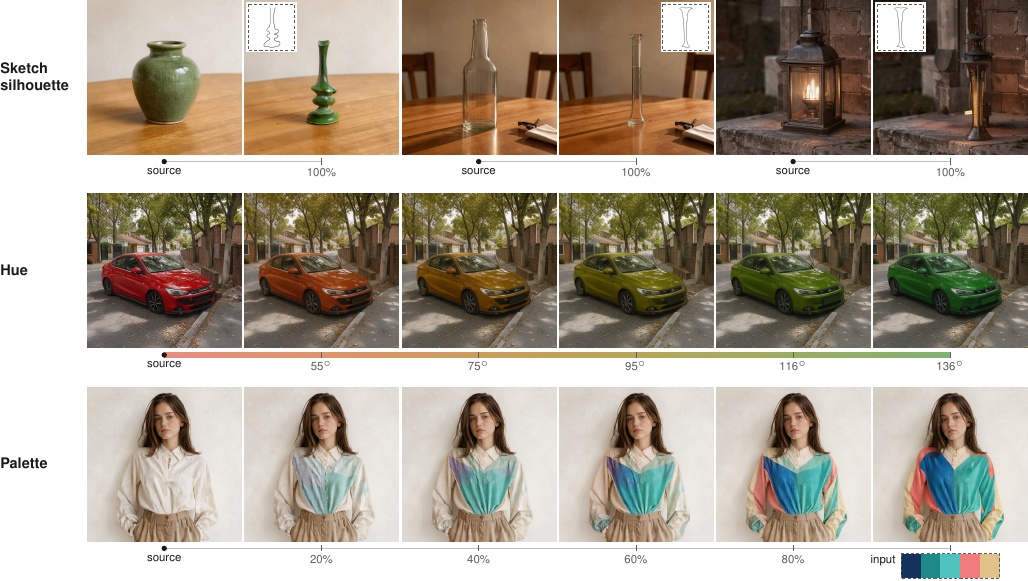}
\caption{Wizmos fitted on FLUX.3~[klein]. Top: Sketch silhouette at the source and at full strength, with the input
sketch in the corner. Hue turns the car from its own hue to green (CIELAB hue angle); Palette moves the blouse to the input palette.}
\label{fig:sliders_flux3}
\end{figure*}

\paragraph{Which fitting images to use.}
\Cref{tab:ablation_sources} varies the number of seeds and prompt
wordings per object while keeping the optimization schedule fixed.
For Size, distributing a fixed source budget across prompt wordings
improves test accuracy over using more seeds of one wording. Palette
benefits from additional seeds even with a single wording. Thus the
benefit of a new wording versus another seed depends on the attribute
and evaluation split.

\subsection{Fitting on a second generator}
\label{sec:supp_second_generator}

Object transfer reuses a fitted mapping. To test whether the
\emph{procedure} carries over to another architecture, we fit new Hue,
Palette and Sketch silhouette wizmos on FLUX.3~[klein]. We keep the reference construction,
losses, offset network and bounds from \klein{}. We apply the offsets
at steps~2--4 (Sketch silhouette: 1--4) in the fifth single-stream block, matching the number of
blocks in which the image and text tokens have interacted. Each fit
uses nine sources (three objects and three prompts per object; for Sketch
silhouette, three vase prompts with three seeds each),
$768^2$ images and four denoising steps; Palette uses one target
palette and Sketch silhouette one sketch. \Cref{fig:sliders_flux3} shows Hue and Palette sweeps on a fitting
source and Sketch silhouette endpoints on new prompts.
The quantitative transfer tests use \klein{}.

\subsection{Cost of fitting and reuse}
\label{sec:appendix_cost}

The one-dimensional fits in \cref{tab:fitting} take 0.9--15.7 GPU-hours on an H100 (median 5.4), with 12--37\,GB
of peak memory. The shared mapping and its $3072\times4$ output layer
per active step require 17k learned parameters for one step and 54k for four,
or 66--210\,KB in 32-bit floats.

Once the source generation has supplied the cached normalization
described in \cref{sec:supp_fit}, each new parameter value takes one
four-step denoising pass, including when several wizmos act together.
The following rendering times exclude that one-time source pass. On an H100 at $768^2$,
rendering takes 0.264\,s per frame without wizmos and 0.265\,s with five
wizmos on five objects. Both use 8.8\,GB of peak memory.

\section{Interactive results on the project page}
\label{sec:supp_examples}

The gallery lets readers explore pre-rendered results across parameter
values, source scenes and combinations of controls. Its views
complement the construction and analysis above.

\paragraph{Exploring individual controls.}
\emph{Drag a wizmo} and \emph{Every wizmo} include scalar sweeps,
a hue wheel, a two-dimensional position plane, and a path from the
source through two sketches. They expose the intermediate changes
behind the selected frames in the paper.
The hue arc and curved position path in \cref{fig:sliders} illustrate
how one fitted mapping follows different paths through its domain.
The illustrated Position and Horizon height examples use a subject
crop as a reference from step~1. Horizon dragging selects the sample
nearest the cursor's measured height; quantitative evaluations use
the fixed parameter grids in \cref{sec:appendix_protocol}.

\paragraph{Comparing and combining controls.}
\emph{Other methods} extends the shared-edit comparisons
(\cref{sec:supp_comparisons,fig:baselines}); the transfer views show
how edits progress on the object sets in
\cref{sec:supp_transfer,fig:generality}. \emph{Two at once} varies
one control along each axis of a grid, holding the other parameter
fixed along each row or column
(\cref{sec:supp_compose,fig:composition}).

\emph{Between inputs} instead blends fitted offsets. Along a triangle
edge, two controls contribute; an interior point combines all three
with barycentric weights $w_k\geq0$ and $\sum_k w_k=1$. In the
colour-wheel exploration of \cref{fig:exploration}, angle selects
the Hue parameter and radius scales its offset. Thus attribute
parameters specify the requested changes, while blending weights
combine or scale the resulting offsets.

\section{Evaluation protocol and measurements}
\label{sec:supp_measurement}

The evaluations ask whether a control reaches its target, progresses
consistently between parameter values, and preserves the rest of the
image. \Cref{sec:appendix_protocol} specifies the sources and
parameter sweeps; \cref{sec:supp_metrics} defines the measurements
and aggregation.

\subsection{Evaluation sources and sweeps}
\label{sec:appendix_protocol}

\paragraph{Sources and splits.}
The quantitative benchmark uses \klein{} sources generated at
$768^2$ in four steps. Test holds out seeds of the fitting prompts;
Transfer uses new prompts and may include new object words. Before
editing, we screen sources in seed order for photographic appearance
(Qwen3-VL) and a single target instance covering at least 4\% of the
image (SAM~3). Hue requires at least 30\% of object pixels to have
CIELAB chroma of 20 or more; ordered attributes require room for the
requested change.

\paragraph{Parameter sweeps.}
A sweep fixes a source and target while varying the parameter over
eleven values. Two-sided controls use $c=-1,-0.8,\ldots,1$;
one-sided controls use $c=0,0.1,\ldots,1$. For Hue, we choose one
of six named colours at least $60^\circ$ from the source hue and
sample eleven angles along the shorter arc, including source and
target. These are passed to the wizmo as absolute hue angles.
The quarter-increment fitting references for Size, horizontal
Position and Horizon height differ from their fifth-increment
evaluation grids, so the sweeps also test interpolation.

The number of source--target sweeps is distinct from the number of
parameter values within each sweep. Per split in \cref{tab:sliders},
there are 27 sweeps for most attributes, nine each for Horizon height
and Ripeness, 30 for Fill level and 36 for Impressionism. Palette
uses 27 sources with two target palettes (54 sweeps); Sketch silhouette
uses nine sources with three target sketches (27 sweeps).

\paragraph{Object-transfer sets.}
\Cref{tab:generality} lists the Original, Near and Far categories.
Original uses new scenes from the fitting categories; Near and Far
use two prompts per category and three seeds per prompt. Categories
that do not yield three usable sources per prompt are excluded before
editing. Hue's Original set covers three of its nine fitting
categories with specified source colours; Near and Far leave colour
open and use chromatic objects. This source selection precedes
application of the fitted controls.

\subsection{Measurements and aggregation}
\label{sec:supp_metrics}

\paragraph{Attribute measurements.}
Measurements follow the attribute units in \cref{tab:sliders}.
For geometry, Size uses object height, Position uses the horizontal
object centre divided by image width, and Sketch silhouette uses
mask IoU with the target outline. These measurements use Grounding
DINO boxes and BiRefNet masks independently of the fitting masks
(boxes enlarged by 5\%, masks thresholded at 0.5). GeoCalib measures
Horizon height as a fraction of image height.

For colour, Hue is the chroma-weighted circular mean over object
pixels with CIELAB chroma of at least 20. Palette is the mean
distance from garment pixels to their nearest swatch, with
$\Delta L^*$ halved. Where SAM~3 is used, we select the target
instance with greatest source-mask overlap.

For ordered levels, Qwen3-VL-8B scores an object crop and we take
the expected value over five verbal anchors. Fill level runs from
empty to full to the brim in approximate quarters. Cartoon style,
Impressionism and Saliency reuse their fitting comparison or
predictor ($^\dagger$): respectively, the reference-comparison loss,
WikiArt probability on the isolated object, and predicted fixation
share. These measure agreement with the fitting criterion and are
reported alongside separately evaluated adherence and identity.

\paragraph{Endpoint and sweep error.}
Let $m_0$, $m$ and $m^\star$ denote source, measured and target values.
PE is $|m-m^\star|$, using angular distance for Hue and
$|\log(m/m^\star)|/\log1.5$ for Size. The parameter-dependent
targets are $m_0\,1.5^c$ for Size, $m_0+0.2c$ for Position, and
$m_0+0.15c$ for Horizon height and Saliency. Hue follows its chosen
angular arc.

The other endpoint targets are zero Palette distance and cartoon-style
loss, unit sketch IoU, an Impressionism probability of 0.9, and
level~5 for Fill level, Ripeness and Weathering (also level~1 at the
negative endpoint for two-sided controls). Their intermediate targets
are $m_0+|c|(m^\star-m_0)$ on the relevant side of the sweep.
PE$_{\mathrm{sweep}}$ averages error over non-source frames, including
endpoints. For two-sided controls, we average the two sides for sweep
error and the two endpoints for PE.

\paragraph{Edit adherence.}
Qwen3.8-27B is used only for evaluation. Scores use normalized
probabilities over the allowed answers with thinking disabled;
pairwise and identity judgments average both image orders.
VQA$_{\mathrm{edit}}$ is the probability of choosing the intended
image from the source and endpoint (chance 0.5). The question is
reversed for a negative endpoint, and visual references accompany
Sketch silhouette and Palette comparisons.

Palette ($^\P$) instead uses a 1--5 agreement rating, from none of
the palette colours to several and no others, scored as
$(\mathbb E[k]-1)/4$. This avoids rewarding saturation alone.
For Position and Horizon height ($^\ddagger$), the judge selects
the line crossing the object centre or horizon from four alternatives:
target, source and two positions displaced by 0.1 of the frame where
possible. We average target probability over label rotations
(chance 0.25).

\paragraph{Progression and evenness.}
Monotonic progression (MP) measures correct ordering over nine pairs
of frames separated by two increments. For Position and Horizon
height, labelled lines at $(-1,-\tfrac13,\tfrac13,1)$ times the
requested change relative to the source locate each frame. We score
the probability that the later frame lies farther in the requested
direction, with half credit for ties. Perceptual evenness (CV) is the
population standard deviation of consecutive DreamSim distances
divided by their mean.

\paragraph{Preservation and image quality.}
For VQA$_{\mathrm{id}}$, the judge ignores the edited attribute and
rates identity from the same object (1) to a different object (5);
we report $(5-\mathbb E[k])/4$. Horizon height uses scene identity,
ignoring camera tilt. Background change ($\Delta$BG) is one minus
mean cosine similarity of DINOv3 ViT-B/16 features outside the source
and endpoint mask union, expanded by 16 pixels. Horizon height has
no background score because it changes the whole frame.
$\Delta$MUSIQ is endpoint minus source quality on the KonIQ-10k
scale (0--100).

\paragraph{Composition and cross-object leakage.}
We combine Size, Hue and cartoon style at their endpoints
(\cref{tab:composition}) and judge each object with the others greyed
out in its crop. VQA$^{\mathrm{all}}_{\mathrm{edit}}$ averages
adherence over intended object--attribute pairs. For an attribute
assigned elsewhere, leakage is $2|p-0.5|$, where $p$ is the probability
of choosing the edited image. We average leakage over unintended pairs
and measure $\Delta$BG outside the edited objects' mask union. Wizmos
act together and fitted LoRAs are merged for one pass. Sequential
editors share a randomized edit order and instruction wording.

\paragraph{Aggregation.}
We average sweeps within each split and wizmos in the All row.
Parameter errors retain their attribute units. Mean adherence excludes
Position, Horizon height and Palette, whose scoring scales differ.

%% file: tab/steps.tex
\begin{table}[t]
\centering
\footnotesize
\setlength{\tabcolsep}{4pt}
\begin{adjustbox}{max width=\linewidth}
\begin{tabular}{@{}lcl@{}}
\toprule
Wizmo & Active steps & Reference or calibration \\
\midrule
\tgroup{3}{Geometry}
Size & 1 & the rough edit for $c$ \\
Position & 1 & the rough edit for $c$ \\
Horizon height & 1 & the source shifted for $c$ \\
Sketch silhouette & 1--4 & the source's loss at step $t$ \\
\midrule
\tgroup{3}{Appearance}
Hue & 2--4 & the source recoloured for $c$ \\
Palette & 2--4 & the source's loss at step $t$ \\
Cartoon style & 2--4 & the source's loss at step $t$ \\
Impressionism & 2--4 & the source's reading at step $t$ \\
\midrule
\tgroup{3}{Perceptual and semantic}
Fill level & 1--4 & the generated reference for $c$ \\
Ripeness & 2--4 & the source's reading at step $t$ \\
Weathering & 2--4 & the source's reading at step $t$ \\
Saliency & 1--4 & the source's reading at step $t$ \\
\bottomrule
\end{tabular}
\end{adjustbox}
\caption{Active denoising steps of the wizmos in \cref{tab:sliders}, out of \klein's four.
A reference image for $c$ is shared across active steps.
Source-relative losses or scores use the unedited clean-image prediction
at the same step as their baseline (\cref{sec:supp_losses}).}
\label{tab:steps}
\end{table}

%% file: tab/fitting.tex
\begin{table}[t]
\centering
\footnotesize
\setlength{\tabcolsep}{3pt}
\begin{adjustbox}{max width=\linewidth}
\begin{tabular}{@{}llrrrrl@{}}
\toprule
Wizmo & Objects & Sources & Iter. & lr & $\rho$ & Preservation \\
\midrule
\tgroup{7}{Geometry}
Size & cat, vase, car & 27 & 200 & 0.12 & 4 & -- \\
Position & cat, vase, car & 27 & 200 & 0.12 & 4 & -- \\
Horizon height & ten sea scenes & 51 & 200 & 0.12 & 1 & -- \\
Sketch silhouette & vase & 9 & 200 & 0.03 & 0.64 & -- \\
\midrule
\tgroup{7}{Appearance}
Hue & nine objects & 81 & 200 & 0.03 & 0.64 & -- \\
Palette & blouse, shirt, sweater & 27 & 200 & 0.03$^\ast$ & 4 & pixels \\
Cartoon style & cat, vase, car & 27 & 200 & 0.03 & 2.56 & -- \\
Impressionism & twelve objects & 36 & 200 & 0.03 & 2.56 & pixels \\
\midrule
\tgroup{7}{Perceptual and semantic}
Fill level & ten vessels & 30 & 240 & 0.12 & 4 & -- \\
Ripeness & banana & 9 & 300 & 0.03 & 1.28 & -- \\
Weathering & bicycle, suitcase, chair & 27 & 300 & 0.03 & 1.28 & -- \\
Saliency & hydrant, lighthouse, chest & 27 & 100 & 0.03 & 0.32 & DINOv3 \\
\bottomrule
\end{tabular}
\end{adjustbox}
\caption{Fitting configurations for the wizmos in \cref{tab:sliders}.
Sources: fitting generations (per target palette or sketch where applicable).
lr is the output-layer learning rate, cosine-decayed to zero ($^\ast$: constant).
$\rho$ bounds the offset's RMS magnitude relative to the source word state.
Preservation compares pixels or DINOv3 features outside the object with
the unedited prediction, using $\mathcal H$ in \cref{eq:loss} with $\lambda=1$.}
\label{tab:fitting}
\end{table}

%% file: tab/sliders.tex
\begin{table*}[t]
\centering
\footnotesize
\setlength{\tabcolsep}{3pt}
\begin{adjustbox}{max width=\linewidth}
\begin{tabular}{@{}llll cc ccc c c c cccc@{}}
\toprule
 & & & & \multicolumn{2}{c}{Edit} & \multicolumn{3}{c}{Sweep} & Local & Identity & Quality & \multicolumn{4}{c}{Transfer} \\
\cmidrule(lr){5-6}\cmidrule(lr){7-9}\cmidrule(l){13-16}
Wizmo & Unit & Fitting signal & Measurement for PE & PE\down & VQA$_\text{edit}$\up & MP\up & CV\down & PE$_\text{sweep}$\down & $\Delta$BG\down & VQA$_\text{id}$\up & $\Delta$MUSIQ\up & PE$_\text{sweep}$\down & VQA$_\text{edit}$\up & VQA$_\text{id}$\up & $\Delta$BG\down \\
\midrule
\tgroup{16}{Geometry}
Size & scale & rough edit & BiRefNet$^\S$ & 0.29 & 0.92 & 0.74 & 0.47 & 0.22 & 0.095 & 0.84 & $-$0.1 & 0.30 & 0.85 & 0.81 & 0.089 \\
Position & share of width & rough edit & BiRefNet$^\S$ & 0.064 & 0.51$^\ddagger$ & 0.63 & 0.43 & 0.049 & 0.121 & 0.80 & 0.1 & 0.057 & 0.44$^\ddagger$ & 0.75 & 0.128 \\
Horizon height & share of height & rough edit & GeoCalib & 0.064 & 0.45$^\ddagger$ & 0.65 & 0.61 & 0.050 & \na & 0.60 & 1.2 & 0.047 & 0.56$^\ddagger$ & 0.67 & \na \\
Sketch silhouette & IoU & SAM 3 & BiRefNet$^\S$ & 0.26 & 1.00 & 0.71 & 0.75 & 0.16 & 0.075 & 0.16 & $-$5.3 & 0.19 & 0.99 & 0.16 & 0.105 \\
\midrule
\tgroup{16}{Appearance}
Hue & degrees & recoloured target & mask hue & 9.6 & 1.00 & 0.78 & 0.57 & 11.4 & 0.007 & 0.92 & $-$1.2 & 17.8 & 0.99 & 0.90 & 0.006 \\
Palette & CIELAB & OT palette loss & CIELAB distance & 11.4 & 0.68$^\P$ & 0.64 & 1.13 & 7.4 & 0.009 & 0.92 & 0.3 & 8.9 & 0.62$^\P$ & 0.86 & 0.007 \\
Cartoon style & share of loss & restyled paste & \na & 0.47$^\dagger$ & 1.00 & 0.74 & 0.68 & 0.24$^\dagger$ & 0.028 & 0.71 & $-$0.7 & 0.31$^\dagger$ & 0.97 & 0.70 & 0.019 \\
Impressionism & style probability & WikiArt classifier & \na & 0.31$^\dagger$ & 0.99 & 0.75 & 1.09 & 0.25$^\dagger$ & 0.009 & 0.49 & $-$4.8 & 0.27$^\dagger$ & 0.99 & 0.47 & 0.009 \\
\midrule
\tgroup{16}{Perceptual and semantic}
Fill level & level & image generator & Qwen3-VL & 0.70 & 0.88 & 0.77 & 0.72 & 0.56 & 0.074 & 0.78 & 1.5 & 0.55 & 0.83 & 0.73 & 0.068 \\
Ripeness & level & Gemma 4 & Qwen3-VL & 0.50 & 0.99 & 0.82 & 1.00 & 0.49 & 0.008 & 0.76 & 2.7 & 0.43 & 0.99 & 0.73 & 0.006 \\
Weathering & level & Gemma 4 & Qwen3-VL & 0.80 & 1.00 & 0.73 & 1.10 & 0.43 & 0.012 & 0.58 & $-$2.1 & 0.53 & 1.00 & 0.60 & 0.015 \\
Saliency & fixation share & DeepGaze IIE & \na & 0.059$^\dagger$ & 0.62 & 0.54 & 0.63 & 0.045$^\dagger$ & 0.014 & 0.67 & 0.0 & 0.065$^\dagger$ & 0.58 & 0.61 & 0.024 \\
\midrule
\textbf{All} &  &  &  & \na & 0.93 & 0.71 & 0.76 & \na & 0.041 & 0.69 & $-$0.7 & \na & 0.91 & 0.67 & 0.043 \\
\bottomrule
\end{tabular}
\end{adjustbox}
\caption{Attribute tracking on new generations. Each row identifies the fitting
signal and the measurement used for endpoint error (PE) and error across
sampled parameter values (PE$_{\mathrm{sweep}}$). Transfer uses new prompts
without refitting. $^\S$: BiRefNet within a Grounding DINO box;
$^\dagger$: PE and PE$_{\mathrm{sweep}}$ reuse the fitting comparison or predictor;
$^\ddagger$: four-line position choice; $^\P$: palette-agreement rating.
The All-row edit score excludes Position, Horizon height and Palette.
MP, CV and PE$_{\mathrm{sweep}}$ measure sweeps; other metrics measure endpoints.}
\label{tab:sliders}
\end{table*}

%% file: tab/generality.tex
\begin{table*}[t]
\centering
\footnotesize
\setlength{\tabcolsep}{4pt}
\begin{subtable}{\textwidth}
\centering
\caption{Transfer after fitting on multiple object categories.}
\label{tab:transfer_multi_category}
\begin{adjustbox}{max width=\linewidth}
\begin{tabular}{@{}lll cc c c@{}}
\toprule
 & & & \multicolumn{2}{c}{Edit} & Local & Identity \\
\cmidrule(lr){4-5}
Wizmo & Tier & Objects & PE\down & VQA$_\text{edit}$\up & $\Delta$BG\down & VQA$_\text{id}$\up \\
\midrule
\multirow{3}{*}{Size} & Original & cat, vase, car & 0.47 & 0.83 & 0.092 & 0.81 \\
 & Near & dog, jug, truck & 0.50 & 0.80 & 0.071 & 0.83 \\
 & Far & shoe, lamp, guitar & 0.68 & 0.72 & 0.074 & 0.86 \\
\cmidrule(l){2-7}
\multirow{3}{*}{Position} & Original & cat, vase, car & 0.063 & 0.48$^\ddagger$ & 0.130 & 0.74 \\
 & Near & dog, jug, truck & 0.076 & 0.42$^\ddagger$ & 0.100 & 0.79 \\
 & Far & shoe, lamp, guitar & 0.137 & 0.31$^\ddagger$ & 0.105 & 0.80 \\
\cmidrule(l){2-7}
\multirow{3}{*}{Fill level} & Original & ten vessels & 0.59 & 0.83 & 0.068 & 0.73 \\
 & Near & vase, bucket & 1.07 & 0.80 & 0.056 & 0.62 \\
 & Far & basket, box & 1.14 & 0.86 & 0.067 & 0.72 \\
\cmidrule(l){2-7}
\multirow{3}{*}{Hue} & Original & ginger cat, green vase, red car (3 of 9) & 14.6 & 1.00 & 0.006 & 0.91 \\
 & Near & dog, jug, taxi & 30.1 & 1.00 & 0.006 & 0.85 \\
 & Far & umbrella, balloon, tent & 14.6 & 1.00 & 0.006 & 0.87 \\
\cmidrule(l){2-7}
\multirow{3}{*}{Palette} & Original & blouse, shirt, sweater & 13.0 & 0.62$^\P$ & 0.007 & 0.83 \\
 & Near & skirt, pants, coat & 15.9 & 0.41$^\P$ & 0.003 & 0.90 \\
 & Far & umbrella, car, chair & 24.5 & 0.26$^\P$ & 0.003 & 0.92 \\
\cmidrule(l){2-7}
\multirow{3}{*}{Cartoon style} & Original & cat, vase, car & 0.48$^\dagger$ & 1.00 & 0.022 & 0.67 \\
 & Near & dog, jug, truck & 0.63$^\dagger$ & 0.90 & 0.016 & 0.73 \\
 & Far & shoe, lamp, guitar & 0.47$^\dagger$ & 0.89 & 0.014 & 0.77 \\
\cmidrule(l){2-7}
\multirow{3}{*}{Impressionism} & Original & twelve household objects & 0.35$^\dagger$ & 0.99 & 0.009 & 0.47 \\
 & Near & clock, table, coat & 0.73$^\dagger$ & 0.96 & 0.007 & 0.62 \\
 & Far & cat, car, house & 0.86$^\dagger$ & 0.76 & 0.007 & 0.68 \\
\cmidrule(l){2-7}
\multirow{3}{*}{Saliency} & Original & hydrant, lighthouse, chest & 0.092$^\dagger$ & 0.58 & 0.024 & 0.61 \\
 & Near & telephone box, windmill, suitcase & 0.119$^\dagger$ & 0.54 & 0.025 & 0.75 \\
 & Far & gramophone, mantel clock, locomotive & 0.130$^\dagger$ & 0.53 & 0.014 & 0.76 \\
\cmidrule(l){2-7}
\multirow{3}{*}{Weathering} & Original & bicycle, suitcase, white chair & 0.86 & 1.00 & 0.016 & 0.59 \\
 & Near & motorcycle, backpack, stool & 1.37 & 0.99 & 0.023 & 0.49 \\
 & Far & book, door & 1.66 & 0.97 & 0.009 & 0.70 \\
\bottomrule
\end{tabular}
\end{adjustbox}
\end{subtable}

\medskip
\begin{subtable}{\textwidth}
\centering
\caption{Generalization stress test after fitting on a single object or scene category.}
\label{tab:transfer_single_category}
\begin{adjustbox}{max width=\linewidth}
\begin{tabular}{@{}lll cc c c@{}}
\toprule
 & & & \multicolumn{2}{c}{Edit} & Local & Identity \\
\cmidrule(lr){4-5}
Wizmo & Tier & Objects & PE\down & VQA$_\text{edit}$\up & $\Delta$BG\down & VQA$_\text{id}$\up \\
\midrule
\multirow{3}{*}{Horizon height} & Original & sea & 0.065 & 0.56$^\ddagger$ & \na & 0.67 \\
 & Near & lake, frozen lake, river & 0.076 & 0.35$^\ddagger$ & \na & 0.64 \\
 & Far & desert, wheat field, snowy plain & 0.075 & 0.39$^\ddagger$ & \na & 0.68 \\
\cmidrule(l){2-7}
\multirow{3}{*}{Sketch silhouette} & Original & vase & 0.32 & 0.99 & 0.105 & 0.16 \\
 & Near & jug, bottle, urn & 0.50 & 0.88 & 0.087 & 0.33 \\
 & Far & lamp, candle, lantern & 0.55 & 0.73 & 0.065 & 0.50 \\
\cmidrule(l){2-7}
\multirow{3}{*}{Ripeness} & Original & banana & 0.50 & 0.99 & 0.006 & 0.73 \\
 & Near & pear, mango, peach & 1.77 & 0.73 & 0.004 & 0.75 \\
 & Far & tomato, strawberry, avocado & 1.82 & 0.53 & 0.005 & 0.86 \\
\midrule
\tgroup{7}{Overall across both panels}
\multirow{3}{*}{\textbf{All}} & Original & & \na & 0.91 & 0.044 & 0.66 \\
 & Near & & \na & 0.84 & 0.036 & 0.69 \\
 & Far & & \na & 0.78 & 0.034 & 0.76 \\
\bottomrule
\end{tabular}
\end{adjustbox}
\end{subtable}
\caption{Analysis of transfer without refitting. Original uses fitting categories
in new scenes; Near and Far use related and more distant categories for which
the attribute remains meaningful. The panels distinguish transfer after fitting
on multiple categories from generalization after fitting on a single category.
The final rows summarize all wizmos across both panels.
Metrics and marks follow \cref{tab:sliders}; \cref{sec:appendix_protocol}
describes sampling. \Cref{fig:generality} illustrates Fill level,
Sketch silhouette and Ripeness.}
\label{tab:generality}
\end{table*}

%% file: tab/composition.tex
\begin{table*}[t]
\centering
\footnotesize
\setlength{\tabcolsep}{3.4pt}
\begin{adjustbox}{max width=\linewidth}
\begin{tabular}{@{}l cc ccc ccc}
\toprule
 & \multicolumn{2}{c}{One object} & \multicolumn{6}{c}{Multiple objects, one edit each} \\
\cmidrule(lr){2-3}\cmidrule(l){4-9}
 & \multicolumn{2}{c}{2 edits} & \multicolumn{3}{c}{2 objects} & \multicolumn{3}{c}{3 objects} \\
\cmidrule(lr){2-3}\cmidrule(lr){4-6}\cmidrule(l){7-9}
Method & VQA$^\text{all}_\text{edit}$\up & $\Delta$BG\down & VQA$^\text{all}_\text{edit}$\up & $\Delta$BG\down & Leak\down & VQA$^\text{all}_\text{edit}$\up & $\Delta$BG\down & Leak\down \\
\midrule
\tgroup{9}{Fitting-based sliders}
Concept Sliders~\cite{gandikota2024concept} & \sworst 0.77 & 0.116 & \sworst 0.73 & 0.116 & 0.52 & \sworst 0.70 & \worst 0.189 & \worst 0.56 \\
Dual-Process~\cite{luo2025dual} & \worst 0.51 & 0.106 & \worst 0.53 & \second 0.091 & \best 0.30 & \worst 0.53 & \second 0.110 & \best 0.28 \\
\midrule
\tgroup{9}{Instruction-based sliders}
SliderEdit~\cite{zarei2026slideredit} & 0.89 & \worst 0.141 & \best 0.89 & 0.121 & 0.48 & \second 0.85 & 0.121 & 0.46 \\
Kontinuous Kontext~\cite{parihar2026kontinuous} & \second 0.90 & \sworst 0.140 & \second 0.88 & \worst 0.150 & \sworst 0.53 & 0.84 & \sworst 0.163 & 0.49 \\
VeloEdit~\cite{li2026veloedit} & 0.89 & 0.132 & \second 0.88 & \sworst 0.128 & \worst 0.54 & \best 0.86 & 0.119 & \sworst 0.53 \\
\midrule
\tgroup{9}{Reference}
\klein editor~\cite{bfl2026flux2klein} & 0.83 & \second 0.095 & 0.86 & 0.108 & \worst 0.54 & 0.84 & 0.150 & 0.50 \\
\midrule
\tgroup{9}{Ours}
Ours & \best 0.91 & \best 0.094 & 0.84 & \best 0.078 & \second 0.36 & \second 0.85 & \best 0.080 & \second 0.29 \\
\bottomrule
\end{tabular}
\end{adjustbox}
\caption{Combining size, hue and cartoon-style controls:
two edits on one object, or one edit per object.
VQA$^{\text{all}}_{\text{edit}}$ averages individual edit scores;
Leak measures responses to edits intended for other objects.
Offsets and LoRAs are applied simultaneously; instruction-based
methods and the reference editor apply edits sequentially.
SliderEdit's joint-instruction variant is not tested.}
\label{tab:composition}
\end{table*}

%% file: tab/comparison_text.tex
\begin{table*}[t]
\centering
\footnotesize
\setlength{\tabcolsep}{3.4pt}
\begin{adjustbox}{max width=\linewidth}
\begin{tabular}{@{}l ccc c cc ccc c c c ccc ccc}
\toprule
 & \multicolumn{4}{c}{Endpoint} & \multicolumn{5}{c}{Sweep} & Local & Identity & Quality & \multicolumn{6}{c}{Transfer} \\
\cmidrule(lr){2-5}\cmidrule(lr){6-10}\cmidrule(l){14-19}
 & \multicolumn{3}{c}{PE\down} & & & & \multicolumn{3}{c}{PE$_\text{sweep}$\down} & & & & \multicolumn{3}{c}{PE$_\text{sweep}$\down} & & & \\
\cmidrule(lr){2-4}\cmidrule(lr){8-10}\cmidrule(lr){14-16}
Method & Size & Hue & Palette & VQA$_\text{edit}$\up & MP\up & CV\down & Size & Hue & Palette & $\Delta$BG\down & VQA$_\text{id}$\up & $\Delta$MUSIQ\up & Size & Hue & Palette & VQA$_\text{edit}$\up & VQA$_\text{id}$\up & $\Delta$BG\down \\
\midrule
\tgroup{19}{Fitting-based sliders}
Concept Sliders~\cite{gandikota2024concept} & 0.91 & 17.8 & 19.0 & 0.76 & 0.65 & 0.79 & 0.55 & 36.3 & \second 9.6 & \sworst 0.075 & \sworst 0.69 & \sworst $-$3.8 & 0.58 & 42.6 & \second 9.6 & 0.75 & \sworst 0.68 & 0.064 \\
Dual-Process~\cite{luo2025dual} & \worst 1.02 & 108.6 & \sworst 24.7 & \worst 0.42 & \worst 0.51 & \best 0.57 & \worst 0.61 & 59.3 & \sworst 13.7 & 0.074 & 0.79 & \second 0.1 & \worst 0.62 & 61.8 & \sworst 12.7 & \worst 0.46 & 0.79 & \sworst 0.066 \\
\midrule
\tgroup{19}{Instruction-based sliders}
SliderEdit~\cite{zarei2026slideredit} & 0.68 & 11.3 & 22.8 & 0.74 & 0.65 & \worst 1.63 & 0.45 & 49.0 & \worst 14.6 & 0.026 & 0.89 & $-$0.2 & 0.47 & 50.3 & \worst 14.7 & 0.74 & \second 0.89 & 0.027 \\
Kontinuous Kontext~\cite{parihar2026kontinuous} & \second 0.61 & 11.0 & \worst 27.5 & 0.71 & 0.66 & 1.12 & 0.45 & 47.8 & 11.0 & 0.043 & 0.86 & $-$3.3 & 0.46 & 53.9 & 10.4 & 0.70 & 0.85 & 0.049 \\
VeloEdit~\cite{li2026veloedit} & 0.69 & \second 10.1 & 20.6 & 0.74 & \best 0.75 & 0.86 & 0.45 & \second 32.6 & 10.4 & 0.035 & 0.86 & \worst $-$4.0 & 0.42 & \second 39.3 & \second 9.6 & 0.74 & 0.85 & 0.036 \\
\midrule
\tgroup{19}{Text-space steering}
Ekin \& Gandelsman~\cite{ekin2026unreasonable} & 0.73 & \worst 116.0 & 24.2 & 0.51 & 0.55 & \second 0.71 & 0.43 & \worst 64.2 & 13.6 & 0.047 & 0.73 & $-$0.3 & 0.46 & \worst 64.2 & \sworst 12.7 & \sworst 0.54 & 0.69 & 0.041 \\
\quad extended search & 0.90 & \sworst 115.9 & 22.1 & \sworst 0.48 & \sworst 0.54 & 0.92 & 0.53 & \sworst 63.9 & 11.8 & 0.054 & 0.70 & $-$0.4 & 0.48 & \sworst 63.9 & 10.9 & \sworst 0.54 & 0.69 & 0.048 \\
Baumann et al.~\cite{baumann2025continuous} & \sworst 0.98 & 60.9 & 23.0 & 0.53 & \sworst 0.54 & \sworst 1.14 & \sworst 0.59 & 53.7 & 13.5 & \second 0.020 & 0.81 & $-$0.2 & \sworst 0.59 & 63.4 & \sworst 12.7 & 0.55 & 0.77 & \second 0.017 \\
\quad no delay & 0.65 & 30.7 & 21.7 & 0.66 & 0.60 & 1.02 & \second 0.38 & 46.1 & 12.1 & \worst 0.081 & \worst 0.58 & \best 0.5 & \second 0.39 & 59.2 & 12.3 & 0.67 & \worst 0.53 & \worst 0.081 \\
\midrule
\tgroup{19}{Reference}
\klein editor~\cite{bfl2026flux2klein} & 0.89 & 19.0 & \second 14.6 & \second 0.83 & \na & \na & \na & \na & \na & \best 0.010 & \best 0.92 & $-$0.2 & \na & \na & \na & \second 0.80 & \best 0.90 & \best 0.008 \\
\midrule
\tgroup{19}{Ours}
Ours & \best 0.29 & \best 9.6 & \best 11.4 & \best 0.87 & \second 0.72 & 0.72 & \best 0.22 & \best 11.4 & \best 7.4 & 0.037 & \second 0.90 & $-$0.3 & \best 0.30 & \best 17.8 & \best 8.9 & \best 0.82 & 0.85 & 0.034 \\
\bottomrule
\end{tabular}
\end{adjustbox}
\caption{The Size, Hue and Palette benchmark from \cref{tab:comparison},
extended with text-space steering. The paired rows vary the strength search
for Ekin \& Gandelsman and the onset of steering for Baumann et al.
Metrics follow the main table. Shading ranks all methods shown here.}
\label{tab:comparison_text}
\end{table*}

%% file: tab/ablation.tex
\begin{table*}[t]
\centering
\footnotesize
\setlength{\tabcolsep}{3.4pt}
\begin{minipage}[c]{0.815\linewidth}
\begin{adjustbox}{max width=\linewidth}
\begin{tabular}{@{}l cc c cccc c c c ccc}
\toprule
 & \multicolumn{3}{c}{Edit} & \multicolumn{4}{c}{Sweep} & Local & Identity & Quality & \multicolumn{3}{c}{Two objects} \\
\cmidrule(lr){2-4}\cmidrule(lr){5-8}\cmidrule(l){12-14}
 & \multicolumn{2}{c}{PE\down} & & & & \multicolumn{2}{c}{PE$_\text{sweep}$\down} & & & & & & \\
\cmidrule(lr){2-3}\cmidrule(lr){7-8}
 & Size & Palette & VQA$_\text{edit}$\up & MP\up & CV\down & Size & Palette & $\Delta$BG\down & VQA$_\text{id}$\up & $\Delta$MUSIQ\up & VQA$_\text{edit}$\up & $\Delta$BG\down & Leak\down \\
\midrule
Full method & \sworst 0.28 & \second 11.4 & \second 0.78 & \second 0.68 & \sworst 0.81 & 0.21 & \second 7.4 & \second 0.055 & \second 0.75 & \worst $-$0.2 & \second 0.86 & \second 0.049 & 0.25 \\
\midrule
\tgroup{14}{Parameter-to-offset mapping}
Scaled offset $c\,u$ & \worst 0.43 & 19.8 & 0.63 & \sworst 0.63 & \worst 0.92 & \worst 0.29 & 12.4 & \best 0.045 & \sworst 0.71 & \worst $-$0.2 & \sworst 0.74 & \best 0.038 & \best 0.14 \\
\midrule
\tgroup{14}{What is fitted}
LoRA, weight scaled by $c$ & \second 0.25 & \best 7.8 & \best 0.88 & \best 0.70 & \second 0.53 & \second 0.20 & \best 6.0 & \worst 0.079 & \worst 0.70 & \best 0.5 & \best 0.88 & \worst 0.084 & \worst 0.77 \\
\midrule
\tgroup{14}{Where the offset is added}
The object's image tokens & \best 0.19 & \sworst 24.4 & \sworst 0.62 & 0.64 & 0.70 & \best 0.15 & \worst 14.3 & 0.058 & \second 0.75 & \second 0.1 & 0.78 & 0.061 & \second 0.18 \\
Every image token & \sworst 0.28 & \worst 25.9 & \worst 0.49 & \worst 0.61 & \best 0.33 & \sworst 0.23 & \sworst 14.2 & \sworst 0.070 & \best 0.78 & \worst $-$0.2 & \worst 0.66 & \sworst 0.070 & \sworst 0.36 \\
\bottomrule
\end{tabular}
\end{adjustbox}
\end{minipage}\hfill
\begin{minipage}[c]{0.175\linewidth}
\centering
\setlength{\tabcolsep}{0.6pt}
\renewcommand{\arraystretch}{0.5}
\begin{tabular}{@{}cc@{}}
\scriptsize Source & \scriptsize LoRA, both only \\
\includegraphics[width=0.49\linewidth]{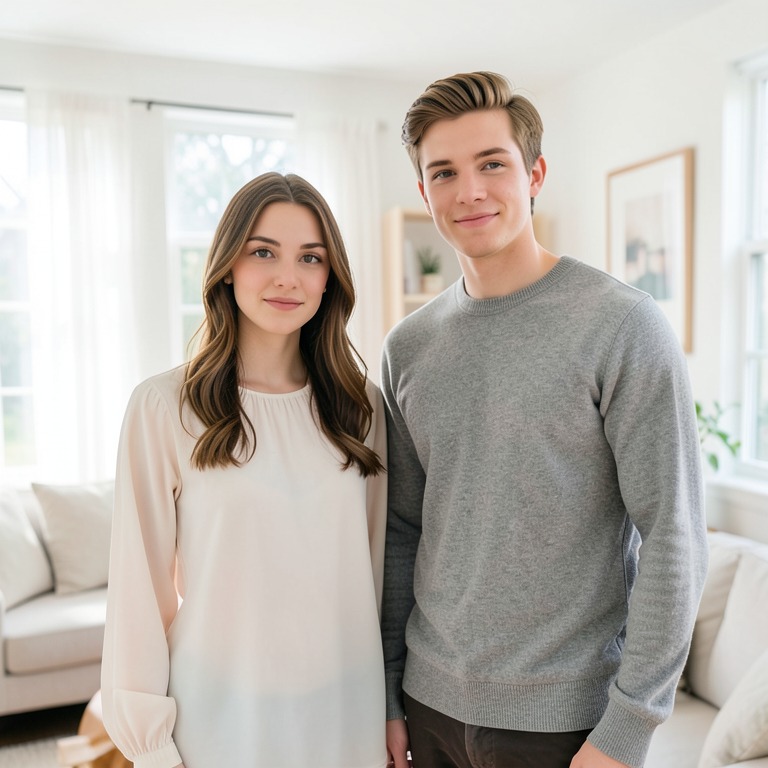} & \includegraphics[width=0.49\linewidth]{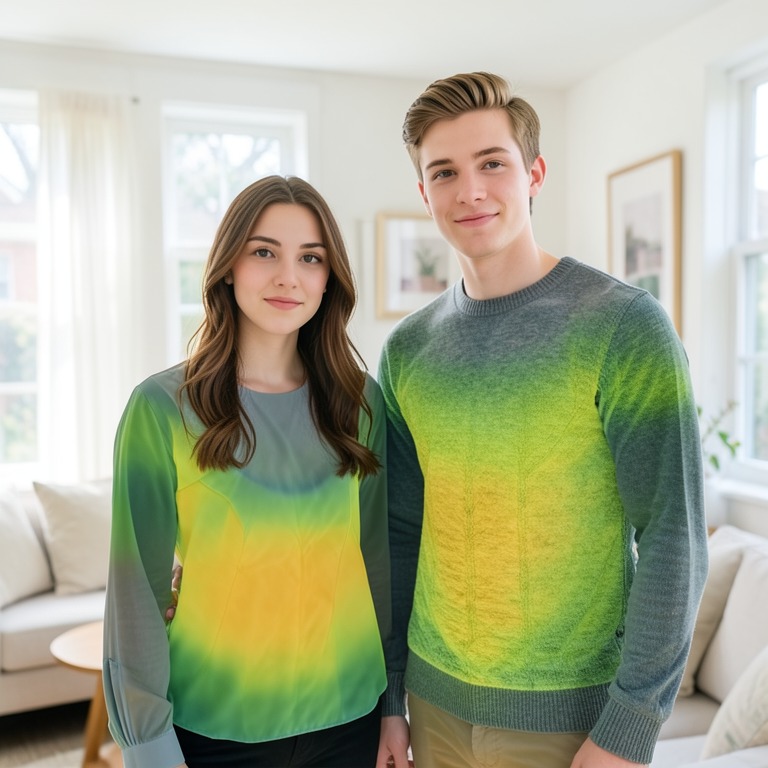} \\[1pt]
\scriptsize Ours, blouse & \scriptsize Ours, sweater \\
\includegraphics[width=0.49\linewidth]{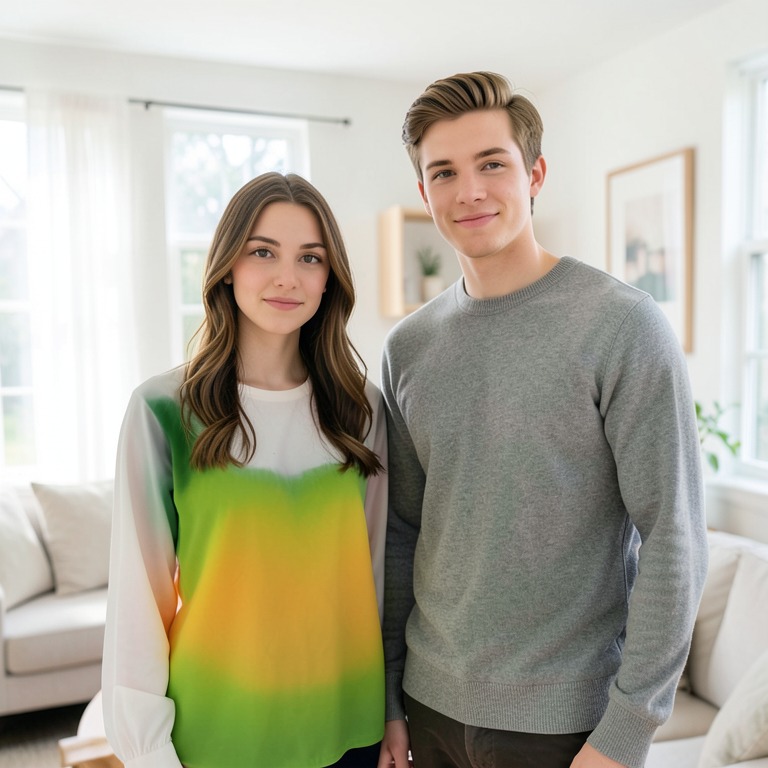} & \includegraphics[width=0.49\linewidth]{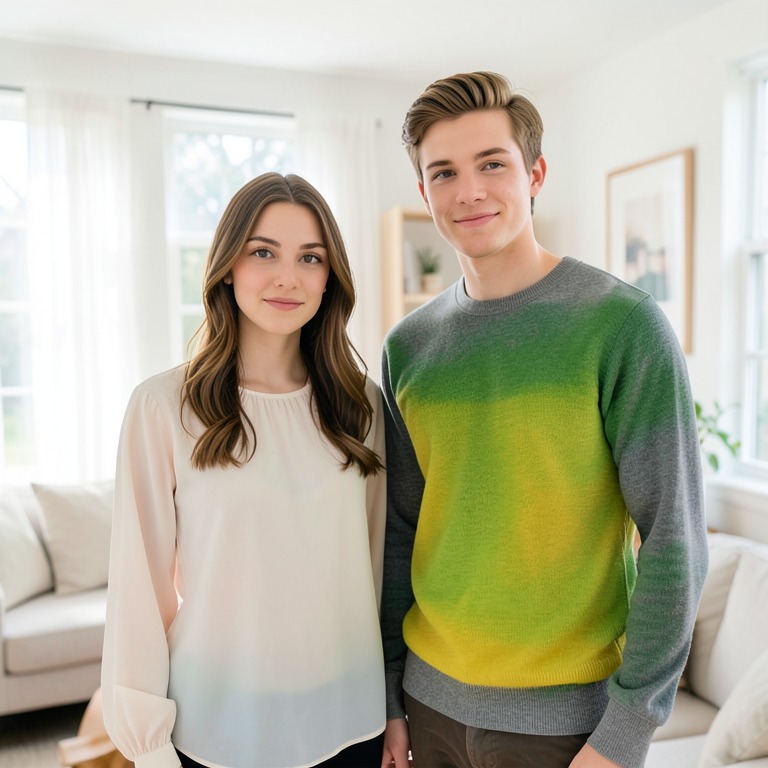} \\
\end{tabular}
\end{minipage}
\caption{Ablations on Size and Palette (\cref{tab:comparison}),
with the same fitting supervision and schedule. Full versus scaled word offsets
tests the parameter mapping; scaled word offsets versus scaled LoRA holds linear
scaling fixed. The image-token variants change the full method's insertion location.
Two-object tests measure edit adherence on the selected object and leakage on the other.
Metrics follow \cref{tab:sliders}. Green and red mark the best and worst results,
with dark and light shades indicating first and second.
Right: our offsets edit the requested garment; the fitted LoRA produces the same
image for either request (sweeps in \cref{fig:ablation}).}
\label{tab:ablation}
\end{table*}

%% file: tab/ablation_sources.tex
\begin{table*}[t]
\centering
\footnotesize
\setlength{\tabcolsep}{3.4pt}
\begin{adjustbox}{max width=\linewidth}
\begin{tabular}{@{}cc cc c cccc c c c cc ccc}
\toprule
\multicolumn{2}{c}{Sources} & \multicolumn{3}{c}{Edit} & \multicolumn{4}{c}{Sweep} & Local & Identity & Quality & \multicolumn{5}{c}{Transfer} \\
\cmidrule(r){1-2}\cmidrule(lr){3-5}\cmidrule(lr){6-9}\cmidrule(l){13-17}
 & & \multicolumn{2}{c}{PE\down} & & & & \multicolumn{2}{c}{PE$_\text{sweep}$\down} & & & & \multicolumn{2}{c}{PE\down} & & & \\
\cmidrule(lr){3-4}\cmidrule(lr){8-9}\cmidrule(lr){13-14}
Seeds & Prompts & Size & Palette & VQA$_\text{edit}$\up & MP\up & CV\down & Size & Palette & $\Delta$BG\down & VQA$_\text{id}$\up & $\Delta$MUSIQ\up & Size & Palette & VQA$_\text{edit}$\up & VQA$_\text{id}$\up & $\Delta$BG\down \\
\midrule
1 & 1 & \worst 0.38 & \worst 14.6 & \worst 0.75 & \worst 0.66 & 0.80 & \worst 0.29 & \worst 9.4 & \worst 0.060 & \worst 0.69 & \second $-$0.2 & \worst 0.45 & \worst 16.0 & \worst 0.71 & \best 0.66 & \worst 0.058 \\
\midrule
3 & 1 & \sworst 0.34 & 13.5 & \sworst 0.76 & \worst 0.66 & 0.80 & \sworst 0.25 & \sworst 9.0 & \second 0.055 & \sworst 0.72 & \sworst $-$0.3 & \worst 0.45 & \sworst 15.8 & \sworst 0.72 & \best 0.66 & 0.053 \\
1 & 3 & \second 0.29 & \sworst 13.9 & 0.77 & \sworst 0.67 & 0.78 & \second 0.22 & 8.8 & \sworst 0.058 & 0.73 & \second $-$0.2 & \best 0.41 & \sworst 15.8 & 0.73 & \second 0.65 & 0.053 \\
\midrule
9 & 1 & \sworst 0.34 & \second 10.6 & \second 0.82 & \best 0.69 & \worst 0.82 & \sworst 0.25 & 7.2 & \worst 0.060 & 0.73 & \second $-$0.2 & \best 0.41 & 13.5 & \second 0.78 & \worst 0.63 & \sworst 0.054 \\
3$^\ast$ & 3 & \best 0.28 & 11.4 & 0.78 & \second 0.68 & \sworst 0.81 & \best 0.21 & 7.4 & \second 0.055 & \best 0.75 & \second $-$0.2 & \second 0.43 & 13.4 & 0.73 & \best 0.66 & \best 0.050 \\
\midrule
15 & 1 & \sworst 0.34 & \best 8.7 & \best 0.88 & \best 0.69 & \second 0.77 & 0.24 & \second 7.1 & \second 0.055 & 0.73 & \best 0.0 & \second 0.43 & \best 10.6 & \best 0.83 & \sworst 0.64 & \sworst 0.054 \\
3 & 5 & 0.30 & \second 10.6 & 0.80 & \best 0.69 & \best 0.76 & \best 0.21 & \best 6.5 & \best 0.054 & \second 0.74 & \worst $-$0.4 & \second 0.43 & \second 12.7 & 0.75 & \sworst 0.64 & \second 0.051 \\
\bottomrule
\end{tabular}
\end{adjustbox}
\caption{Number and variety of fitting sources, on Size and Palette (\cref{tab:ablation}).
Each wizmo is fitted on the given number of seeds of each of the given number of prompts per object
(the prompts are rewordings of one scene), with the full method's schedule,
and evaluated on the test and transfer sources of \cref{tab:sliders}.
Rows with one or three sources per object average three fits on disjoint sources.
$^\ast$: the full method.
Metrics and shading follow \cref{tab:ablation}.}
\label{tab:ablation_sources}
\end{table*}